\documentclass[twoside]{article}

\usepackage[round]{natbib}

\usepackage[accepted]{aistats2019}

\usepackage[utf8]{inputenc} % allow utf-8 input
\usepackage[T1]{fontenc}    % use 8-bit T1 fonts
\usepackage{hyperref}       % hyperlinks
\usepackage{url}            % simple URL typesetting
\usepackage{booktabs}       % professional-quality tables
\usepackage{amsfonts}       % blackboard math symbols
\usepackage{nicefrac}       % compact symbols for 1/2, etc.
\usepackage{microtype}      % microtypography
 
\usepackage{graphicx}

 {\tiny }
\usepackage{algorithm}

\usepackage{xspace}
\usepackage{tikz}
\usetikzlibrary{fit, positioning, shapes,arrows}
\usepackage{amsmath}
\usepackage{amssymb}
\usepackage{float}
\usepackage{multirow}

\usepackage[tableposition=top]{caption}
\usepackage{multicol}
\DeclareCaptionLabelFormat{andtable}{#1~#2  \&  \tablename~\thetable}
\usepackage{xfrac}
\usepackage{wrapfig}
\usepackage{cutwin,lipsum}
\usepackage{picinpar}
\tikzset{three sided/.style={
        draw=none,
        append after command={
            [shorten <= -0.5\pgflinewidth]
            ([shift={(-1.5\pgflinewidth,-0.5\pgflinewidth)}]\tikzlastnode.north east)
        edge([shift={( 0.5\pgflinewidth,-0.5\pgflinewidth)}]\tikzlastnode.north west) 
            ([shift={( 0.5\pgflinewidth,-0.5\pgflinewidth)}]\tikzlastnode.north west)
        edge([shift={( 0.5\pgflinewidth,+0.5\pgflinewidth)}]\tikzlastnode.south west)            
            ([shift={( 0.5\pgflinewidth,+0.5\pgflinewidth)}]\tikzlastnode.south west)
        edge([shift={(-1.0\pgflinewidth,+0.5\pgflinewidth)}]\tikzlastnode.south east)
        }
    }
}

\makeatletter
\newcommand*\bigcdot{\mathpalette\bigcdot@{.5}}
\newcommand*\bigcdot@[2]{\mathbin{\vcenter{\hbox{\scalebox{#2}{$\m@th#1\bullet$}}}}}
\makeatother

\makeatletter
\def\BState{\State\hskip-\ALG@thistlm}
\makeatother
\usepackage{algpseudocode}
\algnewcommand{\Inputs}[1]{%
	\State \textbf{Inputs:} 
}
\algnewcommand{\Initialize}[1]{%
	\State \textbf{Initialize:}
}
\algnewcommand{\Output}[1]{%
	\State \textbf{Output:}
}

\algnewcommand{\prediction}[1]{%
	\State \textbf{}
}

\usepackage[export]{adjustbox}
\makeatletter
\def\algbackskip{\hskip-\ALG@thistlm}
\makeatother

\usepackage{dsfont}

\usepackage{enumitem}
\usepackage{cleveref}
\usepackage{textcomp}
\usepackage{subcaption,siunitx,booktabs} % STUFF INCLUDED BY US (i.e. VA) --> CHECK DOES NOT VIOLATE GUIDELINES

\newcommand{\ie}{i.e.\xspace}
\newcommand{\eg}{e.g.\xspace}

\newcommand{\eq}{Eq.\xspace}

\newcommand{\fig}{Fig.\xspace}
\newcommand{\secref}[1]{\S \ref{#1}}

\newcommand{\tbl}{Tab.\xspace}

\newcommand{\acro}[1]{\textsc{#1}\xspace}
\newcommand{\lgcp}{\acro{lgcp}}
\newcommand{\icm}{\acro{icm}}
\newcommand{\lgcpn}{\acro{mcpm}}
\newcommand{\lgcpnnormal}{\acro{mcpm-n}}
\newcommand{\lgcpngp}{\acro{mcpm-gp}}
\newcommand{\mlgcp}{\acro{mlgcp}}

\newcommand{\gptext}{\acro{gp}}

\newcommand{\mcmc}{\acro{mcmc}}

\newcommand{\lcm}{\acro{lcm}}

\newcommand{\btb}{\acro{btb}}
\newcommand{\gt}{\acro{gt}}
\newcommand{\mv}{\acro{mv}}

\newcommand{\nyc}{\acro{nyc}}
\newcommand{\uk}{\acro{uk}}
\newcommand{\nlpl}{\acro{nlpl}}
\newcommand{\rmse}{\acro{rmse}}
\newcommand{\crime}{\acro{crime}}
\newcommand{\cpu}{\acro{cpu}}
\newcommand{\gb}{\acro{gb}}
\newcommand{\ram}{\acro{ram}}
\newcommand{\ci}{\acro{ci}}
\newcommand{\mtpp}{\acro{mtpp}}

\newcommand{\termkl}{\acro{kl}}
\newcommand{\termelbo}{\acro{elbo}}
\newcommand{\mgf}{\acro{mgf}}
\newcommand{\elltext}{\acro{ell}}
\newcommand{\lemma}{\acro{lemma}}

\newcommand{\coverage}{\acro{ec}}
\newcommand{\sinone}{\acro{s1}}
\newcommand{\sintwo}{\acro{s2}}

\newcommand{\mat}[1]{\mathbf{#1}}
\renewcommand{\vec}[1]{ \mathbf{#1} } % math bold
\newcommand{\vecS}[1]{\boldsymbol{ #1 }  } % this for boldsymbols

\newcommand{\gp}{\mathcal{GP}}
\newcommand{\kernel}{\kappa}
\newcommand{\meanfunc}[1]{m(#1)}
\newcommand{\covfunc}[3]{\kernel(#1,#2; #3)}
\newcommand{\hyperparam}{\vectheta}

\newcommand{\dataset}{\mathcal{D}}
\newcommand{\x}{\vec{x}}

\newcommand{\xprime}{\x^{\prime}}
\newcommand{\vectheta}{\vecS{\theta}}
\newcommand{\y}{\vec{y}}

\newcommand{\f}{\mat{F}}

\newcommand{\fq}{\f_{\bullet q}}

\newcommand{\fn}{\f_{n \bullet}}

\newcommand{\W}{\mat{W}}
\renewcommand{\wp}{\W_{p \bullet}}

\newcommand{\wq}{\W_{\bullet q}}

\newcommand{\Y}{\mat{Y}}

\renewcommand{\u}{\mat{U}}
\newcommand{\uq}{\u_{\bullet q}}

\newcommand{\Z}{\mat{Z}}
\newcommand{\Zq}{\Z_q}

\newcommand{\meanfuqn}{\tilde{\mu}_{nq}}  % conditonal prior mean
\newcommand{\covfuqnn}{\widetilde{K}^{qn}}  % conditonal prior covariance
\newcommand{\likeparam}{\phi}

\newcommand{\priormeanwqp}{\gamma_{pq}}

\newcommand{\Kq}{\mat{K}_{xx}^q}

\newcommand{\covwq}{\vecS{\mat{K}}_{w}^q}
\newcommand{\covwqpp}{K_{w}^{qp}}

\newcommand{\Kzzq}{\mat{K}_{zz}^q}
\newcommand{\Kxzq}{\mat{K}_{xz}^q}
\newcommand{\Kzxq}{\mat{K}_{zx}^q}
\newcommand{\Kzzqinv}{(\mat{K}_{zz}^q)^{-1}}

\newcommand{\poisson}{\text{Poisson}}
\newcommand{\normal}{\mathcal{N}}

\newcommand{\kl}[2]{\mathrm{KL}(#1 \lVert #2)}
\newcommand{\expectation}[2]{ \mathbb{E}_{#1}{\left[#2\right]} }
\renewcommand{\det}[1]{\left\lvert#1\right\rvert}
\newcommand{\trace}{\mbox{ \rm tr }}
\newcommand{\bigO}{\mathcal{O}}

\newcommand{\varparam}{\vecS{\nu}}
\newcommand{\varmeanuq}{\vec{m}_q}

\newcommand{\varcovuq}{\mat{S}_q}
\newcommand{\varmeanwq}{\vecS{\omega}_q}
\newcommand{\varmeanwpq}{\omega_{pq}}
\newcommand{\varcovwq}{\vecS{\Omega}_q}

\newcommand{\calL}{\mathcal{L}}
\newcommand{\elbo}{\calL_{\text{elbo}}}
\newcommand{\ellterm}{\calL_{\text{ell}}}
\newcommand{\klterm}{\calL_{\text{kl}}}
\newcommand{\enterm}{\calL_{\text{ent}}}
\newcommand{\crossterm}{\calL_{\text{cross}}}
  
\newcommand{\qfn}{q(\fn) }
\newcommand{\qwp}{q(\wp)}

\title{Log Gaussian Cox Process Networks}

\begin{document}
\twocolumn[
\aistatstitle{
	Efficient Inference in Multi-task Cox Process Models}
\aistatsauthor{ Virginia Aglietti \And Theodoros Damoulas \And  Edwin V. Bonilla}

\aistatsaddress{ University of Warwick \\ The Alan Turing Institute \And  University of Warwick \\ The Alan Turing Institute   \And CSIRO's Data61 \\ UNSW } 
]

\begin{abstract}
We generalize the log Gaussian Cox process (\lgcp) framework to model multiple correlated point data jointly. The observations are treated as realizations of multiple \lgcp{s}, whose log intensities are given by linear combinations of latent functions drawn from Gaussian process priors. The combination coefficients are also drawn from Gaussian processes and can incorporate additional dependencies. We derive closed-form expressions for the moments of the intensity functions and develop an efficient variational inference algorithm that is orders of magnitude faster than competing deterministic and stochastic approximations of multivariate \lgcp{s}, coregionalization models, and multi-task permanental processes. Our approach  outperforms these benchmarks in multiple problems, offering the current state of the art in modeling multivariate point processes.
 
\end{abstract}

\section{Introduction}
%  POINT PROCESSES ARE IMPORTANT IN REAL-LIFE PROBLEMS
%Many problems in urban science and computational sustainability \citep{gomes2009computational} are characterized by count or point data observed in a spatio-temporal region. Species observations in ecological applications, events that unfold in our cities, crime or other social activities, traffic and human population dynamics are some examples. These processes are inhomogeneous, \ie their underlying intensity varies across space and time. Furthermore, in many settings these processes  do not occur in isolation but can be strongly correlated. For example, in a city such as New York (\nyc), burglaries in different regions can be highly predictive of other crimes' occurrences  such as robberies  and larcenies of motor vehicles. We refer to these settings as multi-task problems and our goal is to exploit such dependencies in order to improve the generalization capabilities of our learning algorithms. 

% EB: Brining back problem description: Problem oriented >> technique oriented
Many problems in urban science and geostatistics are characterized by count or point data observed in a spatio-temporal region. Crime events, traffic or human population dynamics are  some examples. Furthermore, in many settings, these processes can be strongly correlated. For example, in a city such as New York (\nyc), burglaries can be highly predictive of other crimes' occurrences such as robberies and larcenies. These settings are multi-task problems and our goal is to exploit such dependencies in order to improve the generalization capabilities of our learning algorithms.

% LGCP IS A SOUND MATHEMATICAL FRAMEWORK FOR THESE PROCESES
Typically such settings can be modeled as inhomogeneous processes where a space-time varying underlying intensity determines event occurrences. Among these modeling approaches, the log Gaussian Cox process \citep[\lgcp,][]{moller1998log} is one of the most well-established frameworks, where the  intensity is driven by a Gaussian process prior \citep[\gptext{},][]{williams2006gaussian}. The flexibility of \lgcp comes at the cost of incredibly hard inference challenges due to its doubly-stochastic nature and the notorious scalability issues of \gptext models. The computational problems are exacerbated when considering multiple correlated tasks and, therefore, the development of new approaches and scalable inference algorithms for \lgcp models  remains an active area of research  \citep{diggle2013spatial,flaxman2015fast,taylor2015bayesian,leininger2017bayesian}. 

\begin{figure}[t]
	\begin{center}
		\includegraphics[width=0.49\textwidth]{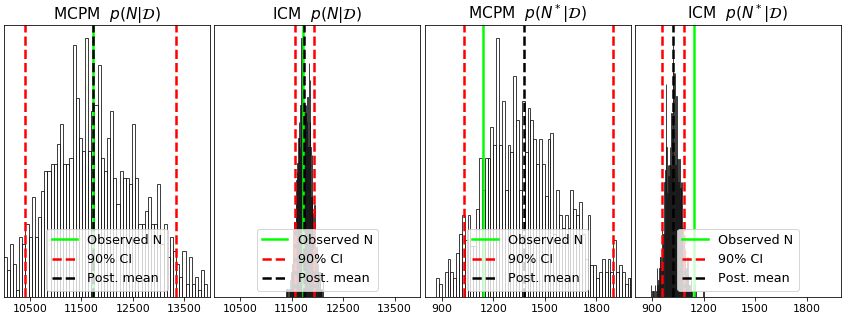}
		\caption{\label{fig:hist_icm_lgcpn} %\crime dataset, \secref{sec:experiments_crime} . %
			%Posterior and predictive distributions of the number of \textit{burglary} events in \nyc  \cite{leininger2017bayesian}. The test set is given in \fig \ref{fig:LGCPN_onCrime_missing}. The solid line represents the ground truth.
			% EB: Avoid refs to sections and figs in the caption. Reviewers want to see everything here
			Posterior and predictive distributions, $p(N | \dataset)$ and $p(N^{*} | \dataset)$ respectively, of the number of burglary events in \nyc using a similar analysis as in \cite{leininger2017bayesian} on the \crime dataset (\secref{sec:experiments_crime}) for our model (\lgcpn) and \icm. The solid line shows the ground truth. Details on the \ci construction are in \secref{sec:experiments}.}
		%Plots for the conditional intensity are given in the supplementary material. 
	\end{center}
\end{figure}

% DEFFICIENCIES OF CURRENT APPROACHES
From a modeling perspective, existing multivariate \lgcp{s} or  linear-coregionalization-model (\lcm) variants for point processes have intensities given by \textit{deterministic} combinations of latent \gptext{s} \citep{diggle2013spatial,taylor2015bayesian, alvarez2012kernels}. These approaches fail to propagate uncertainty in the weights of the linear combination, leading to statistical deficiencies that we address. For instance, \fig \ref{fig:hist_icm_lgcpn} shows how, by propagating uncertainty, our approach (\lgcpn) provides a predictive distribution that contains the counts' ground truth in its 90\% credible interval (\ci). This is not observed for the standard intrinsic coregionalization model \citep[\icm, see e.g.][]{alvarez2012kernels}. 

From an inference point of view, sampling approaches have been proposed  \citep{diggle2013spatial,taylor2015bayesian} and variational  inference algorithms  for models with  \gptext priors and `black-box' likelihoods have been used \citep[see e.g.][]{GPflow2017,dezfouli2015scalable}. While sampling approaches have prohibitive computational cost \citep{shirota2016inference} and mixing issues \citep{diggle2013spatial}, generic methods based on variational inference do not exploit the \lgcp likelihood details and, relying upon Monte Carlo estimates for computing expectations during optimization, can exhibit slow convergence.
  
% CONTRIBUTION 
In this paper we address the modeling and inference limitations of current approaches. More specifically, we make the following contributions.
%\setlist{nolistsep}
%\begin{itemize}[noitemsep,topsep=0ex]
%\item
 
\textbf{Stochastic mixing weights}: We propose a model that considers correlated count data as realizations of multiple \lgcp{s}, where the log intensities are linear combinations of latent \gptext{s} and the combination coefficients are also \gptext{s}.  This provides additional model flexibility and the ability to  propagate uncertainty in a principled way.
%\item 

\textbf{Efficient inference}:  We carry out posterior estimation over  both latent and mixing processes using variational inference. Our method is orders of magnitude faster than competing approaches.
%\item 
 
\textbf{Closed-form  expectations in the variational objective}: We express the required expectations  in the variational inference objective (so called evidence lower bound), in terms of moment generating functions (\mgf{s}) of the log intensities, for which we provide analytical expressions.  We thus avoid Monte Carlo estimates altogether (which are commonplace in modern variational inference methods).   
%\item 

\textbf{First existing experimental comparison and state-of-the-art performance}: To the best of our knowledge, we are the first to provide an experimental comparison between existing multi-task point process methods. This is important as there are currently two dominant approaches (based on the \lgcp or on the permanental process), for which there is little insight on which one performs better. Furthermore, we show that our method provides the best predictive performance on  two large-scale multi-task point process problems with very different spatial cross-correlation structures. 
%We assess \textbf{accuracy}, \textbf{scalability} and \textbf{transfer capabilities} of \lgcpn in the presence of missing data. 
%\item \textbf{Ability of handle missing data} We show the ability of \lgcpn to  thus having the potential for several practical applications, such as under-reporting of rates in epidemiology and criminology.
%\end{itemize}

% Plot histogram

 %$p(N | \dataset)$ and $p(N^{*} | \dataset)$ are obtained by simulating $L$ values from the posterior intensity distribution at $\x$, $\{\lambda^{(l)}(\x)\}_{l=1}^L$, and then computing $\sum_{\x_n } N^{(l)}(\x)$ where $N^{(l)}(\x_n) \sim\text{Poisson}(\lambda^{(l)}(\x))$ and $x_n$ is in the training or test set respectively. The CIs are given by the quantiles of the empirical distributions.

%\input{lgcp}
\section{The \lgcpn model}\label{sec:MTCP-model}
%\paragraph{Background} 
%\subsection{Background}
The log Gaussian Cox process \citep[\lgcp,][]{moller1998log} is an inhomogeneous Poisson process with a stochastic intensity function \citep[see e.g.][]{cox1955some}, where the logarithm of the intensity surface is a \gptext \citep{williams2006gaussian}. Given an input $\x$, a \gptext, denoted as $f(\x) \sim \gp(\meanfunc{\x}, \covfunc{\x}{\xprime}{\hyperparam})$,  is completely specified by its mean function $\meanfunc{\x}$ and its covariance function $\covfunc{\x}{\xprime}{\hyperparam}$ parametrized by the hyper-parameters $\hyperparam$. Conditioned on the realization of the intensity function, the number of points in an area, say $A$, is given by $
y_A | \lambda(\x) \sim \poisson \left(\int_{\x \in A} \lambda(\x)d\x\right)$ with $\lambda(\x) = \text{exp}\{f(\x)\}$.

\subsection{Model formulation}
We consider learning problems where we are given a dataset $\mathcal{D}=\{(\x_n,\y_n)\}_{n=1}^N$ where $\x_n \in \mathbb{R}^D$ represents the input and $\y_n \in \mathbb{R}^P$ gives the event counts for the $P$ tasks. We aim at learning the latent intensity functions and make probabilistic predictions of the event counts. Our modelling approach, which we call \lgcpn , is characterized by $Q$ latent functions which are uncorrelated a priori and are drawn from $Q$ zero-mean \gptext{s}, \ie$f_q | \hyperparam_q \sim \gp(\vec{0}, \covfunc{\x}{\xprime}{\hyperparam_q})$, for $q=1, \ldots, Q$. Hence, the prior over all the $N \times Q$ latent function values $\f$ is:
\begin{equation} \label{eq:prior_f}
%p(\pmb{f}_q|\pmb{\theta}_q) &\sim \mathcal{GP}(\pmb{0},K_{xx}^q), \quad q=1,...,Q \text{,} \\
p(\f | \hyperparam_f) = \prod_{q=1}^Q  p( \fq | \hyperparam_f^q) =  \prod_{q=1}^Q \normal( \fq ; \vec{0}, \Kq) \text{,}
\end{equation}
where $\hyperparam_q$ is the set of hyper-parameters for the $q$-th latent function and $\fq = \{f_{q}(\x_n)\}_{n=1}^N$ denotes the values of  latent function $q$ for the observations $\{\x_n\}_{n=1}^N$. 
We model tasks' correlation by linearly combining the above latent functions with a set of stochastic task-specific mixing weights,  $\W \in \mathbb{R}^{P \times Q}$, determining the contribution of each latent function to the overall \lgcp intensity.  We consider two possible prior distributions for $\W$, an independent prior and a correlated prior.
\paragraph{Prior over weights} We assume weights  drawn from $Q$ zero-mean \gptext{s}:
\begin{equation}
\label{eq:prior-weights-correlated}
p(\W |\hyperparam_w) = \prod_{q=1}^Q  p( \wq | \hyperparam_w^q) = \prod_{q=1}^Q  \normal( \wq ; \vec{0}, \covwq) \text{,}
\end{equation}
where $\wq$ represents the $P$ weights for the $q$-th latent function and $\hyperparam_w$ denotes the hyper-parameters. In the independent scenario, we assume uncorrelated weights across tasks and latent functions by making $\covwq$ diagonal. The observations across tasks are still correlated via the linear mixing of latent random functions. 

%Consider a dataset $\mathcal{D}=\{(\x_n,y_n)\}_{n=1}^N$, where $\x_n$ is a $D$-dimensional vector denoting the centroid of the $n$-th cell on a regular computational grid \citep{diggle2013spatial} and $y_n$ is the number of events  in that cell. For each $n$, the \lgcp observation model can be written as $y_n|f(\x_n) \sim \text{Poisson}\left(\text{exp}\left[f(\x_n)\right]\right)$. \\

\paragraph{Likelihood model} The likelihood of events at locations $\{ \x_{n_p} \}$ under $P$ independent inhomogeneous Poisson processes each with rate function $\lambda_p(\cdot)$ is 
\begin{align*}
 \exp\left[ - \sum_{p=1}^{{P}}\int_{\tau}\lambda_p(\x)d\x\right]\prod_{p=1}^{P}\prod_{n_p=1}^{N_p} \lambda_p(\x_{n_p}) ,
\end{align*}
where %$N_p$ is the \# of events for task $p$ and 
$\tau$ is the observation domain.  %  and $n_p$ denotes the location of the $n$-th event for the $p$-th task. 
Following a common approach we introduce a regular computational grid \citep{diggle2013spatial} on the spatial extent and represent each cell with its centroid. 
Under \lgcpn, the likelihood  of the observed counts $\Y = \{\y_n\}_{n=1}^N$ is defined as: 
\begin{equation}
\label{eq:likelihood}
p(\Y |  \f, \W) % &= \prod_{n=1}^N p(\yn | \lambdan ) \\
%&
= \prod_{n=1}^N \prod_{p=1}^P \text{Poisson}\left(y_{np};\exp\left(\wp \f_{n \bullet} + \likeparam_p\right)\right) \text{,}
\end{equation}
where $y_{np}$ denotes the event counts for the $p$-th task at $\x_n$, $\wp$ represents the $Q$ weights for the $p$-th task, $\f_{n \bullet}$ denotes the $Q$ latent function values corresponding to $\x_n$,  and $\likeparam_p$ indicates the task-specific offset to the log-mean of the Poisson process.  

As in the standard \lgcp model, introducing a \gptext prior poses significant computational challenges during posterior estimation as, na\"{i}vely, inference would be dominated by algebraic operations that are cubic in $N$. To make inference scalable, we follow the inducing-variable approach proposed by \cite{2009variational} and further developed by \cite{bonilla2016generic}. To this end, we augment our prior over the latent functions  in \eq \eqref{eq:prior_f} with $M$ underlying \emph{inducing variables} for each latent process. We denote these $M$  inducing variables for latent process $q$ with $\uq$ and their corresponding \emph{inducing inputs} with the $M \times D$ matrix $\Zq$. We will see that major computational gains are realized when $M \ll N$. Hence, we have that  the prior distributions for the inducing variables and the latent functions are:
\begin{align*}
	p(\u | \hyperparam) &= \prod_{q=1}^Q \normal(\uq;\vec{0}, \Kzzq) %\label{eq:augmented_prior-one}
	\\
	p( \f | \u, \hyperparam) &=  \prod_{q=1}^Q \normal(\Kxzq  \Kzzqinv \uq,  
\Kq - A_q \Kzxq)
%\label{eq:augmented_prior-two}
\end{align*}
where $A_q = \Kxzq \Kzzqinv$, $\u$ is the set of all the inducing variables; the  matrices $\Kq$, $\Kxzq$, $\Kzxq$ and $\Kzzq$ are the covariances  induced by evaluating the corresponding covariance functions at all pairwise rows of the training inputs $\mat{X}$ and the inducing inputs $\Zq$; and $\hyperparam = \{\hyperparam_q \}_{q=1}^Q$ represents the set of hyper-parameters for the $Q$ latent functions. Notice that by integrating out $\u$ from the augmented prior distribution we can recover the initial prior distribution in Eq. \eqref{eq:prior_f} exactly.

\section{Inference}\label{sec:variational-inference}
Our goal is to estimate the posterior distribution over all latent variables given the data, \ie $p(\f, \u, \W | \dataset)$. This posterior is analytically intractable and we  resort to variational inference \citep{jordan1999introduction}.
Variational inference methods entail considering a tractable family of distributions and finding the member of this family that is closest to the true posterior. This is done by minimizing the Kullback-Leiber (\termkl) divergence between the joint approximated posterior and the true joint posterior which is equivalent to maximizing the so-called evidence lower bound, $\mathcal{L}_{\text{elbo}}$.

\subsection{Variational Distributions}
We define our variational distribution as
\begin{align}
	q(\f, \u, \W | \varparam) = p(\f | \u)\prod_{q=1}^Q \underbrace{\normal(\varmeanuq, \varcovuq)}_{q(\uq | \varparam_{u_q})} \prod_{q=1}^Q \underbrace{\normal (\varmeanwq, \varcovwq)}_{q(\wq| \varparam_{w_q})} 
	\label{eq:approx_posterior}
\end{align}
 where $\varparam_u = \{\varmeanuq, \varcovuq\}$ and $\varparam_w = \{\varmeanwq, \varcovwq\}$ are the variational parameters. The choice for this variational distribution, in particular with regards to the incorporation of the conditional prior $p(\f | \u)$, is motivated by the work of \citet{2009variational}, and will yield a decomposition of the  evidence lower bound and ultimately will allow scalability to very large datasets through stochastic optimization. When considering an uncorrelated prior over the weights,  we assume an uncorrelated posterior by forcing $\varcovwq$ to be diagonal. 
\eq \ref{eq:approx_posterior} fully define our approximate posterior. With this, we give details of the variational objective function, \ie $\elbo$, we aim to maximize with respect to $\varparam_u$ and $\varparam_w$. 

\subsection{Evidence Lower Bound}
Following standard variational inference arguments,  it is straightforward to show that evidence lower bounds decomposes as the sum of a KL-divergence term ($\klterm$) between the approximate posterior and the prior, and an expected log likelihood term ($\ellterm$), where the expectation is taken over the approximate posterior. We can write $\elbo(\varparam) =  \klterm(\varparam) + \ellterm(\varparam)$ where $\klterm(\varparam)= - \kl{q(\f,\u,\W| \varparam)}{p(\f, \u, \W)}$ and $\ellterm(\varparam) =  \expectation{q(\f,\u,\W | \varparam)}{\log p(\Y | \f, \W)}$.
 
 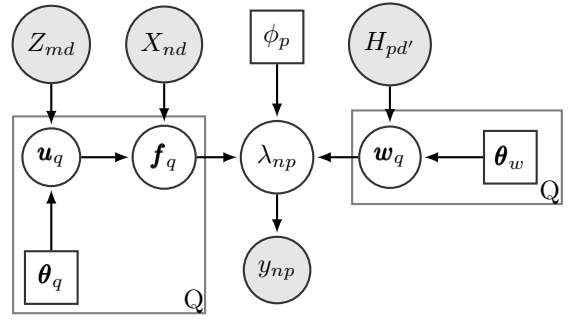
\begin{figure} 
 	\centering
 	\begin{tikzpicture}[shorten >=1pt, node distance=1cm, on grid, auto,thick,scale=0.2]
 
 	\tikzstyle{main}=[circle, minimum size = 7mm, thick, draw =black!80, node distance = 15mm]
 	\tikzstyle{constant}=[rectangle, minimum size = 7mm, draw =black!80, node distance = 16mm, inner sep=0.0mm]
 	\tikzstyle{connect}=[-latex, thick]

 	\node[main] (inducing_process) {$\pmb{u}_{q}$};
 	\node[constant] (theta_q) [below=of inducing_process, label=below:$$] {$\pmb{\theta}_q$};
 	\node[main,fill = black!10] (inducing_inputs) [above=of inducing_process,label=below:$$] {$Z_{md}$};
 	\node[main] (latent_process) [right=of inducing_process,label=below:$$] {$\pmb{f}_{q}$};
 	\node[main, fill = black!10] (inputs) [above=of latent_process,label=below:$$] {$X_{nd}$};
 	\node[main] (intensity) [right=of latent_process,label=below:] {$\lambda_{np}$ };
 	\node[main] (weights) [right=of intensity,label=below:] {$\pmb{w}_{q}$ };
 	\node[constant] (theta_w) [right=of weights,label=below:] {$\pmb{\theta}_w$ };
 	\node[main, fill = black!10] (inputs_gp) [above=of weights,label=below:] {$H_{pd'}$ };
 	\node[main, fill = black!10] (output) [below=of intensity,label=below:] {$y_{np}$ };
 	\node[constant] (offset) [above=of intensity,label=below:] {$\phi_{p}$ };
 	
 	\node[rectangle, inner sep=1.1mm,draw=black!50, fit = (inducing_process) (latent_process) (theta_q)] {};
 	\node[rectangle, inner sep=0mm, fit= (inducing_process) (latent_process) (theta_q),label=below right: Q, xshift=-3.2mm, yshift=3.2mm] {};
 		
 	\node[rectangle, inner sep=2mm,draw=black!50, fit =(weights) (theta_w), xshift = 1.2mm ] {};
 	\node[rectangle, inner sep=0mm, fit= (weights) (theta_w), label=below right: Q, xshift=6.5mm, yshift=2.5mm] {};
 	%\node[rectangle, inner sep=5mm,draw=black!100, dashed, fit =(weights) (theta_w), xshift = 2mm ] {};
 	%\node[rectangle, inner sep=0mm, fit= (weights) (theta_w), label=below right: P, xshift=2.5mm, yshift=-8.5mm] {};
 	 
 	% this is the main border
 	%\node[rectangle, inner sep=6mm,draw=black!20, fit = (theta_q) (inducing_process) (inducing_inputs) (inputs) (intensity) (weights) (output) (offset) (theta_w)] {};
 	
 	\path (theta_q) edge [connect] (inducing_process)
 	(inducing_process) edge [connect] (latent_process)
 	(latent_process) edge [connect] (intensity)
 	(weights) edge [connect] (intensity)
 	(inducing_inputs) edge [connect] (inducing_process)
 	(inputs) edge [connect] (latent_process)
 	(intensity) edge [connect] (output)
 	(theta_w) edge [connect] (weights)
 	(inputs_gp) edge [connect] (weights)
 	(offset) edge [connect] (intensity);
 	\end{tikzpicture}
 	\caption{Graphical model representation of \lgcpn with \gptext prior on $\W$ and tasks' descriptors $H_{pd'}$. Square nodes denote optimised deterministic variables. 
 	} 
 	\label{fig:plates}
 \end{figure}
\paragraph{KL-Divergence Term}
%\subsubsection{KL-Divergence Term}
The variational distribution given in \eq \eqref{eq:approx_posterior} significantly simplifies the computation of $\klterm$, where the terms containing the latent functions $\f$ vanish, yielding $\klterm(\varparam) = \enterm^{u}(\varparam_{u}) +  \crossterm^{u}(\varparam_{u}) + \enterm^{w}(\varparam_{w}) + \crossterm^{w}(\varparam_{w} )$ where each term is given by:
\begin{align*}
\enterm^{u}(\varparam_{u})  &=  \frac{1}{2} \sum_{q=1}^Q \left[ M \log  2\pi + \log \det{\varcovuq} + M\right] \text{,}\\
\crossterm^{u}(\varparam_{u}) &=  \sum_{q=1}^Q  \left[ \log \normal( \varmeanuq; \vec{0}, \Kzzq) - \frac{1}{2} \trace (\Kzzq)^{-1} \varcovuq \right] \text{,}\\
\enterm^{w}(\varparam_{w}) &= \frac{1}{2} \sum_{q=1}^Q \left[\ P \log 2\pi +\log  \det{  \varcovwq} + P\right] \text{,}\\
\crossterm^{w}(\varparam_{w}) &= \sum_{q=1}^Q  \left[ \log \normal(\varmeanwq; \vec{0}, \covwq) - \frac{1}{2} \trace (\covwq)^{-1} \varcovwq \right] \text{.}
\end{align*}
When placing an independent prior and approximate posterior over $\W$, the terms  $\enterm^{w}$ and $\crossterm^{w}$ get simplified further, reducing the computational cost significantly when $P$ is large, see the supplement for the full derivations of the above equations.

\subsection{Closed-form moment generating function of log intensities} 
\lgcpn allows to compute the moments of the intensity function in closed form. The $t$-th moment for the $p$-th task intensity evaluated at $\x_n$, namely $\mathbb{E}\left[\lambda_p(\x_n)^t\right]$, can be written as $ \text{exp}(t\likeparam_p)\mathbb{E}\left[\text{exp}\left(t\wp\f_{n \bullet}\right)\right] = \text{exp}(t\likeparam_p)\mgf_{\wp\f_{n \bullet}}(t)$ where $\mgf_{\wp\f_{n \bullet}}(t)$ denotes the moment generating function of $\wp\f_{n \bullet}$ in $t$. The random variable $\wp\f_{n \bullet}$ is the sum of products of independent Gaussians \citep{craig} and its \mgf is thus given by:
\begin{equation}
\text{MGF}_{\wp\f_{n \bullet}}(t) =  \prod_{q=1}^Q \frac{\text{exp}\left[\frac{t\priormeanwqp\meanfuqn + \frac{1}{2}(\meanfuqn^2\covwqpp + \priormeanwqp^2\covfuqnn)t^2}{1 - t^2 \covwqpp \covfuqnn}\right]}{\sqrt{1-t^2\covwqpp\covfuqnn}},
\label{eq:mgf}
\end{equation}
where the expectation is computed with respect to the prior distribution of $\wp$ and $\f_{n \bullet}$;  $\priormeanwqp$ is the prior mean of $w_{pq}$; $\covfuqnn$ denotes the variance of $f_{nq}$; and $\covwqpp$ is the variance of $w_{pq}$.

\subsection{Closed-form Expected Log Likelihood Term}
%\subsubsection{Closed form Expected Log Likelihood Term
Despite the additional model complexity introduced by the stochastic nature of the mixing weights, the expected log-likelihood term $\ellterm$ can be evaluated in closed form:
\begin{align*}
\ellterm(\varparam)&= - \sum_{n=1}^N \sum_{p=1}^P\text{exp}(\likeparam_p) \underbrace{\mathbb{E}_{\qfn \qwp}\left(\text{exp}\left(\wp\f_{n \bullet}\right)\right)}_{\text{MGF}_{\wp\f_{n \bullet}}(1)} \\ &+\sum_{n=1}^N \sum_{p=1}^P \sum_{q=1}^{Q} \left[y_{np}(\varmeanwpq\mu_{nq} + \likeparam_p) - \text{log}(y_{np}!)\right] 
%\numberthis \label{eq:ell_exact}
\end{align*}
where $\mu_{nq} = \mu_{q}(x^{(n)})= A_q\varmeanuq(x^{(n)})$. The term $\text{MGF}_{\wp\f_{n \bullet}}(1)$ is computed evaluating \eq \eqref{eq:mgf} at $t=1$ given the current variational parameters for $q(\W)$ and $q(\f)$. 
A closed form expected log-likelihood term significantly speeds up the algorithm achieving similar performance but 2 times faster than a Monte Carlo approximation on the \crime dataset (\secref{sec:experiments_crime}, see Fig. (2) and Fig. (3) in the appendix). In addition, by providing an analytical expression for $\ellterm$, we avoid high-variance gradient estimates which are often an issue in modern variational inference method relying on Monte Carlo estimates. 

\textbf{Algorithm complexity and implementation}
The time complexity of our algorithm is dominated by algebraic operations on $\Kzzq$, which are  $\bigO(M^3)$,  while the space complexity is dominated by storing $\Kzzq$, which is  $\bigO(M^2)$ where $M$ denotes the number of inducing variables per latent process.  $\enterm$ and $\crossterm$  only depends on distributions over $M$ dimensional variables thus their computational complexity is independent of $N$. $\ellterm$ decomposes as a sum of expectations over individual data points thus stochastic optimization techniques can be used to evaluate this term making it independent of $N$. 
Finally, the algorithm complexity does not depend on the number of tasks $P$ but only on $Q$ thus making it scalable to large multi-task datasets. We provide an implementation of the algorithm that uses Tensorflow \citep{abadi2016tensorflow}. Pseudocode can be found in the supplementary material.
\section{Related work}\label{sec:related_work}
Our approach relates to other work on multi-task regression, models with black-box likelihoods and \gptext priors, and other \gptext-modulated Poisson processes. 

\textbf{Multi-task regression}: 
A large proportion of the literature on multi-task learning methods \citep{caruana1998multitask} with \gptext{s} has focused on regression problems \citep{teh-et-al-aistats-05,bonilla2008multi,alvarez2010efficient,alvarez2011computationally,wilson-et-al-icml-12,gal-et-al-nips-2014}, perhaps due to the additional challenges posed by complex non-linear non-Gaussian likelihood models. Some of these methods are reviewed in \citet{alvarez2012kernels}. Of particular interest to this paper is the linear coregionalization model (\lcm) of which the \icm is a particular instance. It can be shown that the \lgcpn prior covariance generalizes the \lcm prior. In addition, the two methods differ substantially in terms of inference, model flexibility and accuracy (see \S3 in the supplement). Unlike standard coregionalization methods, \cite{schmidt2003bayesian} consider a prior over the mixing weights but, unlike our method, their focus is on regression problems and they carry out posterior estimation via a costly  \mcmc  procedure. 

\textbf{Black-box likelihood methods}:
Modern advances in variational inference have allowed the development of generic methods for inference in  models with \gptext priors and `black-box' likelihoods including \lgcp{s} \citep{GPflow2017,dezfouli2015scalable,hensman-et-al-aistats-2015}.
While these frameworks offer the opportunity to prototype new models quickly, they can only handle deterministic weights and are inefficient. In contrast, we exploit our likelihood characteristics and derive closed-form \mgf expressions for the evaluation of the \elltext term. By adjusting the \termelbo to include the entropy and cross entropy terms arising from the stochastic weights and using the closed form \mgf, we significantly improve the algorithm convergence and efficiency.
\begin{figure}[t]
	\centering
	\includegraphics[width=0.39\textwidth]{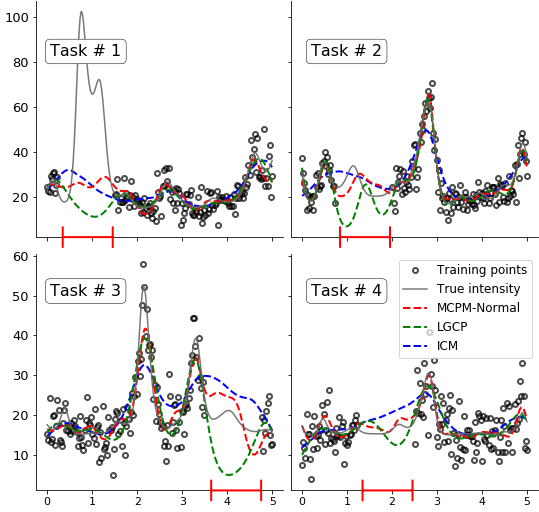}
	\caption{Four related tasks evaluated at 200 evenly spaced points in $[-1.5,1.5]$. |-| denote the 50 contiguous observations removed from the training set of each task.}
	\label{fig:GPprior_synthetic}
\end{figure}
% Table for synthetic
\begin{table}
		\caption{Performance on the missing intervals. \lgcpnnormal and \lgcpngp denote independent and correlated prior respectively. Time in seconds per epoch.} 
		\begin{center}
			\resizebox{0.46\textwidth}{!}{\begin{tabular}{lcccccccccc}
					\toprule
					\multirow{2}{*}{} &
					\multicolumn{4}{c}{\rmse}  & \multicolumn{4}{c}{\nlpl} & \multicolumn{1}{c}{\multirow{2}{*}{\cpu}} \\
					\cmidrule(r){2-5} \cmidrule(r){6-9}
					& \textbf{1} & \textbf{2} & \textbf{3} & \textbf{4} & \textbf{1} & \textbf{2} & \textbf{3} & \textbf{4} & time\\
					\midrule
					{\lgcpnnormal} 	 & 38.61 &  7.86 &  5.71 &  \textbf{4.68} & 20.99 &  3.75 &  \textbf{3.31} &  \textbf{3.02} & \textbf{0.18}\\
					{\lgcpngp} 		& \textbf{38.58} & \textbf{7.69} &  \textbf{5.70} &  4.71 	 & \textbf{20.95} &  \textbf{3.70} &  \textbf{3.31} &  3.03 & 0.25\\
					{\lgcp} & 48.17 & 14.32 & 11.83 &  5.38 & 43.40 &  8.78 &  8.98 &  3.27 & 0.32\\
					{\icm}   & 39.07 &  7.96 &  7.88 &  6.03  &    21.81 &  3.76 &  3.77 &  3.38 & 0.52\\
					\bottomrule
			\end{tabular}}
		\end{center}
	\label{tab:all_synthetic_performance}
\end{table} 

\textbf{Other GP-modulated Poisson processes}:
Rather than using a \gptext prior over the log intensity,  different transformations  of the latent \gptext{s} have been considered as alternatives to model point data. For example,  in the Permanental process,  a \gptext prior is used over the squared root of the intensities \citep{pmlr-v70-walder17a,lloyd-icml-2015,pmlr-v37-lian15, lloyd2016latent, john2018large}.  Similarly,  a sigmoidal transformation of the latent \gptext{s} was studied by   \citet{adams2009tractable} and used in conjunction with convolution processes by \citet{gunter2014efficient}. 
Permanental and Sigmoidal Cox processes are very different from \lgcp/\lgcpn both in terms of statistical and computational properties. There is no conclusive evidence in the literature on which model provides a better characterisation of point processes and under what conditions. The \lgcpn likelihood introduces computational issues in terms of \elltext evaluation which in this work are solved by offering a closed form \mgf function. On the contrary, permanental processes suffer from important identifiability issues such as reflection invariance and, together with sigmoidal processes, do not allow for a closed form prediction of the intensity function.

Among the inhomogeneous Cox process models, the only two multi-task frameworks are \cite{gunter2014efficient} and \citet[\mtpp,][]{pmlr-v37-lian15}. The former suffers from high computational cost due to the use of expensive \mcmc schemes scaling with $\mathcal{O}(PN^3)$. In addition, while the framework is develop for an arbitrary number of latent functions, a single latent function is used in all of the presented experiments. 
\mtpp restricts the input space to be unidimensional and does not handle missing data. Furthermore, none of these two methods can handle spatial segregation  (\btb experiment) through a shared global mean function or a single latent function.

%Synthetic exps
\section{Experiments}\label{sec:experiments}
\begin{figure}[t]
	\centering
	\includegraphics[width=0.39\textwidth]{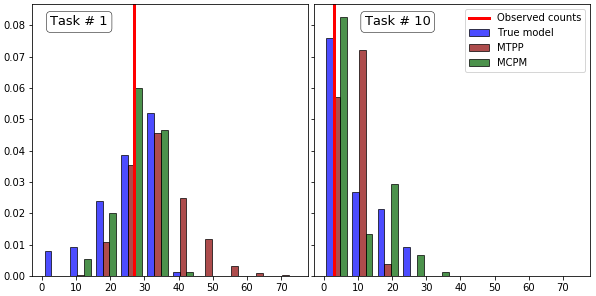}
	\caption{\sintwo dataset. Predicted empirical distribution of event counts for two tasks.}
	\label{fig:synthetic_mtpp}
\end{figure}

We first analyse \lgcpn on two synthetic datasets. In the first one, we illustrate the transfer capabilitites in a missing data setting comparing against \lgcp and \icm. In the second one, we assess the predictive performance against the \mtpp model which cannot handle missing data. We then proceed to model two real world datasets that exhibit very different correlation structures. The first one includes spatially segregated tasks while the second one is characterized by strong positive correlation between tasks. 

\textbf{Baselines}
We offer results on both complete and incomplete data settings while comparing against \mlgcp \citep{taylor2015bayesian}, a \textit{variational} \lgcp model \citep{nguyen2014automated} and a \textit{variational} formulation of \icm with Poisson likelihood implemented in GPflow \citep{GPflow2017, hensman-et-al-aistats-2015}.

\textbf{Performance measures}
We compare the algorithms evaluating the root mean square error (\rmse), the negative log predicted likelihood (\nlpl) and the empirical coverage (\coverage) of the posterior predictive counts distribution. \rmse  and \nlpl values for the $p$-th task are computed as $\rmse_p = \sqrt{\frac{1}{N}\sum_{n=1}^{N}\left(y_{np}-\mathbb{E}(\lambda_{np})\right)^2}$ and $\nlpl_p = - \frac{1}{S}\sum_{s=1}^{S}\frac{\sum_{n=1}^{N}\log p(y_{np}|\lambda_{np}^s)}{n}$ where $\mathbb{E}(\lambda_{np})$ represents the posterior mean estimate for the $p$-th intensity at  $\x_n$ and $S$ denotes the number of samples from the variational distributions $q(\f)$ and $q(\W)$.  The \coverage is constructed by drawing random subregions from the training (in-sample) or the test set (out-of-sample) and evaluating the coverage of the 90\% \ci of the posterior ($p(N | \dataset)$) and predictive ($p(N^{*} | \dataset)$) counts distribution for each subregion $B$ \citep{leininger2017bayesian}. These are in turn obtained by simulating from $N^{(l)}(B) \sim\text{Poisson}(\lambda^{(l)}(B))$ for $l=1,...,L$ with $\lambda^{(l)}(B)$ denoting the $l$-th sample from the intensity posterior and predictive distribution. 
The presented results consider $L = 100$ and consistent results were found when changing this value. In terms of subregions selection, we fix their size, say $Z$, and randomly selected $L$ of them among all the possible areas of size $Z$ in the training or test set.
$\coverage=1$ when all \ci{s} contain the ground truth. Finally, in order to assess transfer in the 2D experiments, we partition the spatial extent in $Z$ subregions and create missing data ``folds'' by combining non-overlapping regions, one for each task. We repeat the experiment $Z$ times until each task's overall spatial extend is covered  thus accounting for areas of both high and low intensity.
 %The \lgcpn performance and computational gains thus stem from our main contributions: the network framework, the random coupled weights and the closed-form \mgf . 
 \footnote{Code and data for all the experiments is provided at \texttt{https://github.com/VirgiAgl/MCPM}}

\begin{table}
		\caption{ \sintwo dataset. Performance on the test intervals. Time in seconds per epoch.} 
		\begin{center}
			\resizebox{0.46\textwidth}{!}{\begin{tabular}{lccccccccccc}
					\toprule
					\multirow{3}{*}{} &
					\multicolumn{10}{c}{\nlpl } & \multicolumn{1}{c}{\multirow{2}{*}{\cpu}}\\
					\cmidrule(r){2-11}
					& \textbf{1} & \textbf{2} & \textbf{3} & \textbf{4} & \textbf{5} & \textbf{6} & \textbf{7} & \textbf{8} & \textbf{9} & \textbf{10} & time\\
					\midrule
					{\lgcpnnormal} & \textbf{1.47} & \textbf{1.46} & \textbf{0.95} & 0.17 & 1.30 &
					1.39 & 1.52 & 0.70 & 1.58 & 0.58  & \textbf{0.03}\\
					{\lgcpngp} & 1.52 & 1.80  & 0.96 & \textbf{0.13} & 1.29 & \textbf{1.37} & 1.61 & \textbf{0.65} & \textbf{1.50} & 0.76 & \textbf{0.03}\\
					{\mtpp} &  1.60 & 3.05 & 1.13 & 0.15 & \textbf{1.24} &
					1.44 & \textbf{1.49} & 1.13 & 1.70 & \textbf{0.52} & 5.97\\
					
					\bottomrule
			\end{tabular}}
		\end{center}

\label{tab:mtpp_lgcpn_nlpl}
\end{table}

\textbf{Synthetic missing data experiment (\sinone)}
To illustrate the \textit{transfer} capabilities of \lgcpn we construct four correlated tasks by sampling from a multivariate point process with final intensities obtained as the linear combination of two latent functions via task-specific mixing weights (\fig \ref{fig:GPprior_synthetic}). The final count observations are obtained by adding noise to the Poisson counts generated through the constructed intensities. When using a coupled prior over the weights, we consider covariates describing tasks (\eg minimum and maximum values) as inputs. \lgcpn is able to reconstruct the task intensities in the missing data regions by learning the inter-task correlation and transferring information across tasks. Importantly, it significantly outperforms competing approaches for all tasks in terms of \coverage (see supplement) and \nlpl. In addition, it has a lower \rmse for $\sfrac{3}{4}$ of the tasks (\tbl \ref{tab:all_synthetic_performance}) while being $\approx$ 3 times faster than \icm.

\textbf{Synthetic comparison to \mtpp (\sintwo)}
To assess the \textit{predictive} capabilities of \lgcpn against \mtpp, which cannot handle missing data, we replicate the synthetic example proposed by \cite{pmlr-v37-lian15} (\textsection6.1). 
We train the models with the observations in the interval $[0,80]$ and predict in the interval $[80,100]$. 
 %Notice that we are comparing a discretized (\lgcpn) versus a continuous (\mtpp) Poisson likelihood. The estimated \mtpp intensities are thus not affected by the discretization error and should better capture sudden intensity changes. 
 \fig \ref{fig:synthetic_mtpp} shows how \lgcpn better recovers the true model event counts distribution with respect to \mtpp. 
 We found \lgcpn to outperform \mtpp in terms of \nlpl, \coverage and \rmse for $\sfrac{7}{10}$ tasks,  see Tab \ref{tab:mtpp_lgcpn_nlpl} , Fig \ref{fig:synthetic_mtpp} and the supplementary material. 
% The current \tf implementation of \lgcpn runs \approx 200 times faster than the Matlab \mtpp implementation.
 
% Tab for crime
%\begin{table}[ht]
%	\caption{\crime dataset. Performance on the missing regions. Time in seconds per epoch.}
%	% These are the results on the land cells
%	%  3.66 hours LGCPN
%	% 38.48 hours ICM
%	
%	\label{tab:performance_crime_missing}
%	\begin{center}
%		

\begin{table}
	\centering
	\caption{\textit{Upper:} \crime dataset. Performance on the missing regions. Time in seconds per epoch. Lower values of \nlpl are better. \textit{Lower:} In-sample/Out-of-sample 90\% \ci coverage for the predicted event counts distributions. Higher values of \coverage are better.} 
	\begin{subtable}{0.46\textwidth}
		\sisetup{table-format=-1.2}   % 2 decimals, leave space for minus sign
		\begin{center}
			\resizebox{1.0\textwidth}{!}{\begin{tabular}{lccccccccccccccc}
					\toprule
					\multirow{2}{*}{} &
					\multicolumn{7}{c}{Standardized \nlpl (per cell)} &  \multicolumn{1}{c}{\multirow{2}{*}{\cpu} } \\
					\cmidrule(r){2-8}
					& \textbf{1} & \textbf{2} & \textbf{3} & \textbf{4} & \textbf{5} & \textbf{6} & \textbf{7} & time\\
					\midrule
					
					\multirow{2}{*}{\lgcpnnormal}& \textbf{0.56} &  0.91 &   \textbf{0.66} &   \textbf{1.09} &
					0.85 &  \textbf{10.29} &   \textbf{0.42}  &  \multirow{2}{*}{ \textbf{2.85}}\\
					&  (0.10) & (0.27) & (0.30) &  (0.27) &  (0.52) &
					(2.51) &  (0.05) \\
					
					\multirow{2}{*}{\lgcpngp}  &  0.72 &   \textbf{0.75} &   0.94 &   1.53 &
					\textbf{0.57} &  18.76 &   0.58  & \multirow{2}{*}{3.11}\\
					& (0.18) & (0.18) & (0.55) &  (0.52) &  (0.19) &
					(8.25) & (0.12) \\
					
					\multirow{2}{*}{\lgcp}  &   9.90 &   9.32 &  19.34 &  5.30 &
					18.18 & 36.73 &   9.68 & \multirow{2}{*}{2.87} \\
					& (3.66) &   (2.41) & (11.45) &   (1.02) &
					(8.65) &  (4.02) &  (2.67)\\
					
					\multirow{2}{*}{\icm}  &  0.87 &  1.36 &  0.91 &  1.19 &
					0.69 &  12.30 &   0.93 & \multirow{2}{*}{44.13}\\
					& (0.27) &  (0.35) & (0.45) & (0.40) & (0.11) &
					(3.02) &  (0.17) \\
					\bottomrule
			\end{tabular}}
		\end{center}
	\end{subtable}
	
	\begin{subtable}{0.46\textwidth}
		\centering
		\sisetup{table-format=4.0} % integer values only, up to 4 digits
		\begin{center}
			\resizebox{1.0\textwidth}{!}{\begin{tabular}{lcccccccc}
					\toprule
					\multirow{3}{*}{} &
					\multicolumn{7}{c}{Empirical Coverage (\coverage) } \\
					\cmidrule(r){2-8}
					& \textbf{1} & \textbf{2} & \textbf{3} & \textbf{4} & \textbf{5} & \textbf{6} & \textbf{7} \\
					\midrule
					{\lgcpnnormal} & 0.99/0.80 & \textbf{1.00}/0.73  & 0.97/\textbf{0.71} & \textbf{1.00}/0.73 & 0.98/0.61 & \textbf{1.00}/\textbf{1.00} & 0.99/\textbf{0.87} \\  
					{\lgcpngp} & \textbf{1.00}/\textbf{0.87} & \textbf{1.00}/\textbf{0.74} & \textbf{1.00}/\textbf{0.71} & \textbf{1.00}/\textbf{0.95} & \textbf{1.00}/\textbf{0.88} & 0.80/\textbf{1.00}   & \textbf{1.00}/0.85\\
					{\lgcp} & 0.86/0.29  & 0.76/0.20 & 0.86/0.29 & 0.82/0.37 & 0.68/0.25 & 0.94/0.00 & 0.83/0.21\\
					{\icm}   & 0.68/0.73 & 0.75/0.50 & 0.64/0.52 & 0.79/0.65  & 0.59/0.78 & 0.93/0.86 &  0.841/0.64 \\
					\bottomrule
					\\
			\end{tabular}}
		\end{center}
	\end{subtable}
	\label{tab:performance_crime_missing}
\end{table}

\subsection{Crime events in NYC}\label{sec:experiments_crime}
\begin{figure*}[h]
	\begin{center}
		\includegraphics[width=0.72\textwidth]{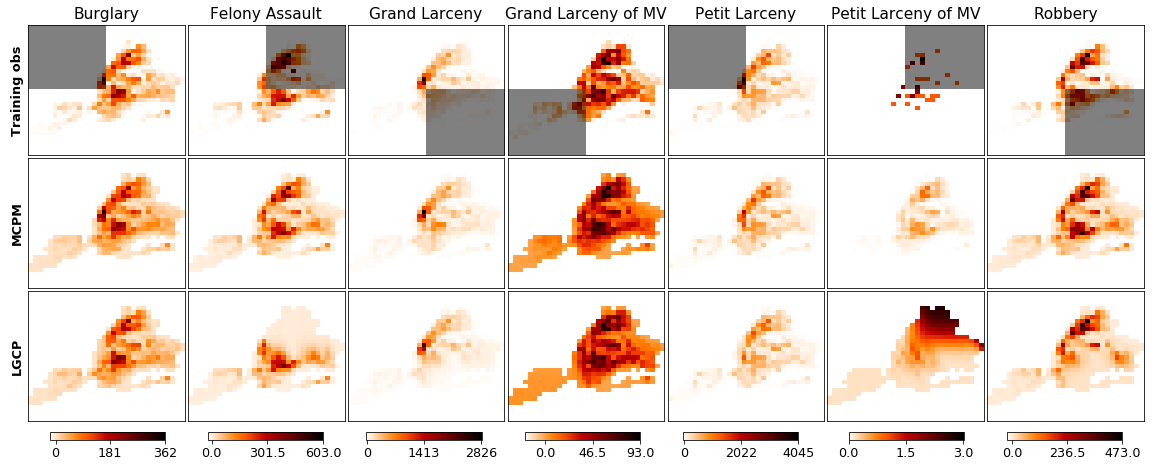}
		\caption{\label{fig:LGCPN_onCrime_missing}\crime dataset. Estimated surface when introducing missing data in the shaded regions. \textit{Row 1:} Counts of crime events on a regular grid. \textit{Row 2:} \lgcpn \textit{Row 3:} \lgcp.}
		%Plots for the conditional intensity are given in the supplementary material. 
	\end{center}
\end{figure*}

In this section we demonstrate the performance, transfer capabilities and scalability of \lgcpn on a real-world dataset recording different crimes in \nyc  (\crime). The dataset includes latitude and longitude locations of burglaries (1), felony assaults (2), grand larcenies (3), grand larcenies of motor vehicle (\mv, 4), petit larcenies (5), petit larcenies of \mv (6) and robberies (7) reported in 2016. The data are discretized into a 32$\times$32 regular grid (Fig.\ref{fig:LGCPN_onCrime_missing}). 
Lack of ground truth intensities typically restricts quantitative measures of generalization and hence we focus on validating and comparing \lgcpn from two different perspectives: i) using complete data so as to assess the quality of the recovered intensities as well as the \textit{computational complexity} and scalability gains over \mlgcp; and  ii) using missing data so as to  validate the \textit{transfer} capabilities of \lgcpn when compared with \lgcp and  \icm.
 %
 % Fig for crime results

\textbf{Complete/Missing Data Experiments} We first consider a full-data experiment and we spatially interpolate the crime surfaces running \lgcpn with four latent functions characterized by Mat\'ern 3/2 kernels. We repeat the experiment with \mlgcp setting the algorithm parameters as suggested by \citet{taylor2015bayesian}. Similar results are obtained with the two methods (see \fig (8) in the supplement for a visualisation of the estimated intensity surfaces). However,  in terms of \textit{computational gains}, an \mlgcp run takes $\approx$ 14 hrs while \lgcpn requires $\approx$ 2 hrs.

To assess \textit{transfer}, we keep the same experimental settings and introduce missing data regions by partitioning the spatial extent in 4 subregions as explained above. 
%into 4 regions. We create missing data ``folds'' by combining 7 non-overlapping regions, one for each task.  We first place the missing region for ``Burglary" in the upper left corner of the grid and we move forward this missing region 4 times until the overall area is covered. 
The shaded regions in Fig. \ref{fig:LGCPN_onCrime_missing} represent one possible configuration of the missing data across tasks. \lgcpn successfully transfers information across tasks (\fig \ref{fig:LGCPN_onCrime_missing}) thereby recovering, for all crime types, the signal in the missing data regions. By exploiting task similarities, the algorithm outperforms competing approaches in all of the tasks, in terms of \coverage, \nlpl (\tbl \ref{tab:performance_crime_missing}) and \rmse (see the supplementary material.) Finally, \lgcpn significantly outruns \icm in terms of algorithm efficiency. \lgcpnnormal converges in 1.19 hrs (1500 epochs) on a Intel Core i7-6t00\acro{u} \cpu (3.40GHz, 8\gb of \ram) while \icm needs 12.26 hrs (1000 epochs).

\begin{table}
	\centering
		\caption{\textit{Upper:} \rmse and \nlpl on \btb with missing data. Time in seconds per epoch. Lower values of \nlpl are better. \textit{Lower:} In-sample/Out-of-sample 90\% \ci coverage for the predicted event counts distributions. Higher values of \coverage are better. } 
	\begin{subtable}{0.46\textwidth}
		\sisetup{table-format=-1.2}   % 2 decimals, leave space for minus sign
		\begin{center}
			\resizebox{1.0\textwidth}{!}{\begin{tabular}{lcccccccccc}
					\toprule
					\multirow{2}{*}{} &
					\multicolumn{4}{c}{\rmse} &
					\multicolumn{4}{c}{\nlpl (per cell)} &  \multicolumn{1}{c}{\multirow{2}{*}{\cpu}}\\
					\cmidrule(r){2-5} \cmidrule(r){6-9}
					& \textbf{\gt 9} & \textbf{\gt 12} & \textbf{\gt 15} & \textbf{\gt 20} &  \textbf{\gt 9} & \textbf{\gt 12} & \textbf{\gt 15} & \textbf{\gt 20}  & time\\
					\midrule
					\multirow{2}{*}{\lgcpnnormal} &  0.83 &  0.24 & 0.28 & 0.29 &   \textbf{1.23} & 0.20 & \textbf{0.33} &   \textbf{0.35} & \multirow{2}{*}{7.73}  \\
					& (0.15) &  (0.07) & (0.07) & (0.10) & (0.40) &  (0.07) &  (0.11) &  (0.16)\\
					
					\multirow{2}{*}{\lgcpngp} &   \textbf{0.81} &   0.22 &  \textbf{0.27} &   \textbf{0.27} & 1.42 & 0.27 &  0.41 & 0.58 & \multirow{2}{*}{ \textbf{7.63}} \\
					& (0.14) &  (0.08) & (0.07) &  (0.09) & (0.42) &  (0.09) &  (0.14) &  (0.24) \\
					
					\multirow{2}{*}{\lgcp }  & 1.37 &  0.61 & 0.63 &  1.24 &    1.70 &  0.48 &  0.72 & 0.86 & \multirow{2}{*}{8.76}  \\
					& (0.33) &  (0.13) & (0.12) & (0.56) & (0.39) & (0.11) & (0.17) & (0.36) \\

					\multirow{2}{*}{\icm  }  &   0.91 &  \textbf{0.21} &  0.32 &  7.24 & 1.44 &  \textbf{0.18} &   0.34 &  0.37 & \multirow{2}{*}{67.06} \\
					& (0.15) &   (0.07) &  (0.08) &  (5.48) & (0.40) &  (0.06) & (0.10) & (0.14) \\
					
					\bottomrule
			\end{tabular}}
		\end{center}
	\end{subtable}
	
	\begin{subtable}{0.46\textwidth}
		\centering
		\sisetup{table-format=4.0} % integer values only, up to 4 digits
\begin{center}
	\resizebox{1.0\textwidth}{!}{\begin{tabular}{lcccccccc}
			\toprule
			\multirow{3}{*}{} &
			\multicolumn{4}{c}{Empirical Coverage (\coverage)} \\
			\cmidrule(r){2-5}
			&\textbf{\gt 9} & \textbf{\gt 12} & \textbf{\gt 15} & \textbf{\gt 20} \\
			\midrule
			{\lgcpnnormal} & 0.87/\textbf{0.92} & 0.97/\textbf{0.99}  & 0.93/0.96 & 0.95/\textbf{1.00} \\  
			{\lgcpngp} & \textbf{0.93}/0.91 & \textbf{0.98}/0.98 & \textbf{0.97}/\textbf{0.98} & \textbf{0.97}/0.99 \\
			{\lgcp} &  0.91/0.79 & 0.97/0.98 & \textbf{0.97}/0.97 & 0.96/0.98 \\
			{\icm} & 0.90/0.84 & 0.96/0.98 & 0.95/0.96 & 0.96/0.96 \\		
			\bottomrule	
	\end{tabular}}
\end{center}
	\end{subtable}
	\label{tab:performance_BTB_missing}
\end{table}

%\begin{table}[t]
%	\caption{Coverage of the 90\% CI for the predicted events counts.Values are reported as (In-sample/Out-of-sample).}
%	\label{tab:coverage_btb}
%	\begin{center}
%		\resizebox{0.49\textwidth}{!}{\begin{tabular}{lcccccccc}
%				\toprule
%				\multirow{3}{*}{} &
%				\multicolumn{4}{c}{Coverage } \\
%				\cmidrule(r){2-5}
%				&\textbf{\gt 9} & \textbf{\gt 12} & \textbf{\gt 15} & \textbf{\gt 20} \\
%				\midrule
%				{\lgcpnnormal} & 0.87/\textbf{0.92} & 0.97/\textbf{0.99}  & 0.93/0.96 & 0.95/\textbf{1.00} \\  
%				{\lgcpngp} & \textbf{0.93}/0.91 & \textbf{0.98}/0.98 & \textbf{0.97}/\textbf{0.98} & \textbf{0.97}/0.99 \\
%				{\icm} & 0.90/0.84 & 0.96/0.98 & 0.95/0.96 & 0.96/0.96 \\
%				
%				\bottomrule
%		\end{tabular}}
%	\end{center}
%\end{table}

%%%%%%%%%%%%%%%%%%%%%%%%%%%%%%

\subsection{Bovine Tuberculosis (\btb) in Cornwall}
\begin{figure}[h]
	\begin{center}
		\includegraphics[width=0.41\textwidth]{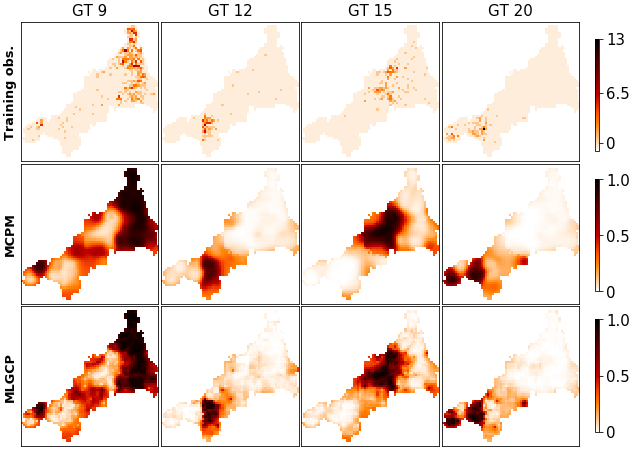}
		\captionof{figure}{Estimated conditional probabilities on \btb (Second row: \lgcpn, Third row:\mlgcp) in the complete data setting. Estimated intensity surfaces are available in the supplementary material.}
		\label{fig:complete_BTB}
		%Plots for the conditional intensity are given in the supplementary material. 
	\end{center}
\end{figure}

%\textbf{Bovine Tuberculosis (\btb) in Cornwall} 
We showcase the performance of \lgcpn on the \btb dataset \citep{diggle2013spatial,taylor2015bayesian} consisting of locations of \btb incidents in Cornwall, \uk (period 1989--2002) and covariates measuring cattle density, see first row in \fig \ref{fig:complete_BTB}. We follow \citet{diggle2013spatial} and only consider the four most common \btb genotypes (\gt: 9, 12, 15 and 20).
 
\textbf{Complete Data Experiments}:
%We run the chain for one million iterations with a burn in of 100,000 iterations and a thinning parameter of 900 iterations. The trace plots for the chain show convergence of the algorithm. 
We estimate the four \btb intensities by fitting an \lgcpn with four latent functions and Mat\'ern 3/2 kernels. We initialise the kernel lenghtscales and variances to 1. For direct comparison, we train the \mlgcp model following the grid size, prior, covariance and \mcmc settings by \citet{taylor2015bayesian}. We run the \mcmc chain for 1M iterations with a burn in of 100K and keep 1K thinned steps. Following \citet{diggle2013spatial}, in \fig \ref{fig:complete_BTB} we report the probability surfaces computed as $\pi_p(x) = \lambda_p(x) / \sum_{p=1}^P \lambda_p(x) $ where $\lambda_p(x)$ is the posterior mean of the intensity for task $p$ at location $x$. Estimated intensities surfaces can be found in the supplementary material. 

The probability surfaces are comparable with both approaches characterizing well the high and low intensities albeit varying at the level of smoothness. In terms of \textit{computational gains} we note that \mlgcp takes $\approx$ 30 hrs for an interpolation run on the four \btb tasks  while \lgcpn only requires $\approx$ 8 hrs. The previously reported \citep{diggle2013spatial} slow mixing and convergence problems of the chain, even after millions of \mcmc iterations, renders \mlgcp problematic for application to large-scale multivariate point processes. Finally, the built-in assumption of a single common \gptext latent process across tasks limits the number and the type of inter-task correlations that we can identify and model efficiently.
%
 % Plot for BTB
 % Plot for BTB
\begin{figure}[h]
	\begin{center}
		\includegraphics[width=0.41\textwidth]{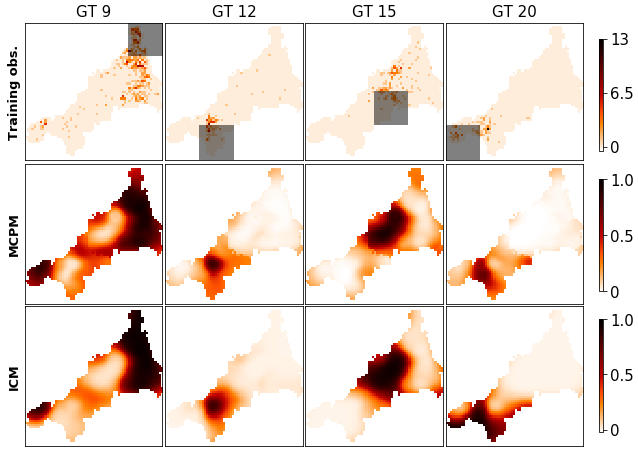}
		\captionof{figure}{\textit{First row:} Counts of the \btb incidents on a 64$\times$64 grid. Shaded areas represent missing data regions. Estimated conditional probabilities by \lgcpn (\textit{second row}) and by \icm (\textit{third row}).}
		\label{fig:transfer_BTB}
		%Plots for the conditional intensity are given in the supplementary material. 
	\end{center}
\end{figure}

\textbf{Missing Data Experiments}:\textit{Transfer} is evaluated by partitioning the space into 16 suregions and constructing missing data regions as explained above. The shaded regions in the first row of \fig \ref{fig:transfer_BTB} represent one such fold of the missing areas across tasks. 
We provide average quantitative metrics across folds for an \lgcpn with four latent functions, Mat\'ern 3/2 kernels and $30\%$ of the training inputs as inducing inputs. As in the complete data setting, we report estimated conditional probabilities in \fig \ref{fig:transfer_BTB}. \lgcpn manages to recover the overall behaviour of the process in the missing regions showing significant transfer of information across spatially segregated tasks while avoiding negative transfer in the case of negative spatial correlation.
 \lgcpn outperforms \lgcp across all tasks and \icm on $\sfrac{3}{4}$ of the tasks (\tbl \ref{tab:performance_BTB_missing}) while achieving the highest \coverage both in-sample and out-of-sample. In addition, \lgcpn has a significant computational advantage: it converges in 3.18 hrs (1500 epochs) while \icm converges in 18.63 hrs (1000 epochs).

\section{Conclusions and Discussion}\label{sec:conclusions}
We proposed a new multi-task learning framework for modeling correlated count data based on \lgcp models. We consider observations on different tasks as being drawn from distinct \lgcp{s} with correlated intensities determined by linearly combined \gptext{s} through task-specific random weights. By considering stochastic weights, we allow for the incorporation of additional dependencies across tasks while providing better uncertainty quantification. We derive closed-form expressions for the moments of the intensity functions and use them to develop an efficient variational algorithm that is order of magnitude faster than the current state of the art. We show how \lgcpn achieves the state of the art performance on both synthetic and real datasets providing a more flexible and up to 15 times faster methodology compared to the available \lgcp multivariate methods. 
Future work will focus on increasing the interpretability of the mixing weights. For instance, placing an alternative sparse prior distribution \citep{titsias2011spike} on $\W$ would induce sparsity and thus act as a model selection mechanism for $Q$. However, introducing a more complex prior on $\W$ would also break the tractability of the variational objective and would require further approximations.  

\subsection*{Acknowledgments}
 This work was supported by the EPSRC grant EP/L016710/1, The Alan
 Turing Institute under EPSRC grant EP/N510129/1 and the Lloyds Register Foundation
 programme on Data Centric Engineering.

\bibliographystyle{apalike}
\bibliography{references}

\end{document}

\twocolumn[
	\aistatstitle{Efficient Inference in Multi-task Cox Process Models \\
		Supplementary Material}
	\aistatsauthor{ Virginia Aglietti \And Theodoros Damoulas \And  Edwin V. Bonilla}
	
	\aistatsaddress{ University of Warwick \\ The Alan Turing Institute \And  University of Warwick \\ The Alan Turing Institute   \And CSIRO's Data61 \\ UNSW } 
	]

\section{Derivation of the KL-divergence Term}\label{sec:derivations}
The \termkl-divergence terms composing the \termelbo can be written as
$\klterm(\varparam) = \enterm^{u}(\varparam_{u}) +  \crossterm^{u}(\varparam_{u}) + \enterm^{w}(\varparam_{w}) + \crossterm^{w}(\varparam_{w} )$ where each term is given by:
\begin{align}
	\crossterm^{u}(\varparam_{u}) &=  \sum_{q=1}^Q  \left[ \log \normal( \varmeanuq; \vec{0}, \Kzzq) - \frac{1}{2} \trace (\Kzzq)^{-1} \varcovuq \right]  \label{eq:crossent_u} \\
	\enterm^{u}(\varparam_{u})  &=  \frac{1}{2} \sum_{q=1}^Q \left[ M \log  2\pi + \log \det{\varcovuq} + M\right] \label{eq:ent_u} \\
	\crossterm^{w}(\varparam_{w}) &= \sum_{q=1}^Q  \left[ \log \normal(\varmeanwq; \vec{0}, \covwq) - \frac{1}{2} \trace (\covwq)^{-1} \varcovwq \right] \label{eq:ent_w}\\
	\enterm^{w}(\varparam_{w}) &= \frac{1}{2} \sum_{q=1}^Q \left[\ P \log 2\pi +\log  \det{  \varcovwq} + P\right] \label{eq:crossent_w}\text{,}
\end{align}
When placing an independent prior and approximate posterior over $\W$, the terms  $\enterm^{w}$ and $\crossterm^{w}$ get simplified further, reducing the computational cost significantly when a large number of tasks is considered. Here we derive the expressions for \eqs \eqref{eq:ent_u}--\eqref{eq:crossent_w}.

The cross-entropy term for $\u$ (\eq \eqref{eq:crossent_u})  is given by:
\begin{align*}
\crossterm^{u}(\varparam_{u}) &=  \expectation{q(\u | \varparam_u)}{\log p(\u)} \\
&=  \int q(\u|\varparam_u) \log p(\u) d \u \\
&= \sum_{q=1}^Q \int q(\uq|\varparam_u) \log p(\uq) d \uq \\
&= \sum_{q=1}^Q \left[  \normal(\uq ; \varmeanuq, \varcovuq) \log \normal(\uq;\vec{0}, \Kzzq) \right] \\
&= \sum_{q=1}^Q  \left[ \log \normal( \varmeanuq; \vec{0}, \Kzzq) - \frac{1}{2} \trace (\Kzzq)^{-1} \varcovuq \right]\text{.}
\end{align*}

The entropy term for $\u$ (\eq \eqref{eq:ent_u}) is given by:
\begin{align*}
\enterm^u(\varparam_{u}) &= - \expectation{q(\u | \varparam_u)}{\log q(\u|\varparam_u)} \\
&= - \int q(\u|\varparam_u) \log q(\u |\varparam_u) d \u \\
&=  - \sum_{q=1}^Q \int q(\uq|\varparam_u) \log q(\uq|\varparam_u)d \uq \\
&= - \sum_{q=1}^Q \int  \normal(\uq ; \varmeanuq, \varcovuq)  \log \normal(\uq ; \varmeanuq, \varcovuq) d \uq \\
&= - \sum_{q=1}^Q \left[ \normal(\varmeanuq; \varmeanuq, \varcovuq) - \frac{1}{2} \trace (\varcovuq)^{-1} \varcovuq \right]\\
&= \frac{1}{2} \sum_{q=1}^Q \left[ M\log 2\pi + \log \det \varcovuq + M \right]\text{.}
\end{align*}

When placing a coupled prior on the mixing weights, the cross-entropy term for $\pmb{W}$  (\eq \eqref{eq:ent_w}) is given by:
\begin{align*}
\crossterm^{w}(\varparam_{w}) &=  \expectation{q(\W | \varparam_w)}{\log p(\W)} \\
&= \int q(\W|\varparam_w) \log p(\W) d \W \\
&= \sum_{q=1}^Q \int q(\wq|\varparam_w) \log p(\wq) d \wq \\
&= \sum_{q=1}^Q \int  \normal(\varmeanwq, \varcovwq) \log \normal(\vec{0}, \covwq)) d \wq \\
&= \sum_{q=1}^Q \left[\log \normal (\varmeanwq ; \vec{0},\covwq) - \frac{1}{2} \trace (\covwq)^{-1} \varcovwq \right]\text{.}
\end{align*}

The entropy term for $\pmb{W}$ (\eq \eqref{eq:crossent_w}) is given by:
\begin{align*}
\enterm^{w}(\varparam_{w}) &=  - \int q(\W|\varparam_{w}) \log q(\W|\varparam_{w}) d \W\\
&= - \sum_{q=1}^Q \int \normal (\wq ; \varmeanwq, \varcovwq) \text{log} \normal (\wq ; \varmeanwq, \varcovwq) d \wq \\
&= - \sum_{q=1}^Q \left[\normal (\varmeanwq ; \varmeanwq,\varcovwq) - \frac{1}{2} \trace (\varcovwq)^{-1} \varcovwq \right]\\
&= \frac{1}{2} \sum_{q=1}^Q \left[ P\log 2\pi + \log \det \varcovwq + P \right]\text{.}
\end{align*}

When placing an independent prior and approximate posterior over $\W$, the terms  $\enterm^{w}$ and $\crossterm^{w}$ get further simplified in:

\begin{align*}
\enterm^{w}(\varparam_{w}) &=  - \int q(\W|\varparam_{w}) \log q(\W|\varparam_{w}) d \W\\
&= - \sum_{q=1}^Q \sum_{p=1}^P  \int \normal ( \varmeanwpq, \varcovwpq) \text{log} \normal (\varmeanwpq, \varcovwpq) d w_{pq}\\
&= \frac{1}{2} \sum_{q=1}^Q \sum_{p=1}^P \left[ \log 2\pi + \log \varcovwpq + 1 \right]\text{,}
\end{align*}

and in:
\begin{align*}
\crossterm^{w}(\varparam_{w}) &=  \int q(\W|\varparam_w) \log p(\W) d \W \\
&= \sum_{q=1}^Q \sum_{p=1}^P \int q(w_{pq}|\varparam_w) \log p(w_{pq}) d w_{pq} \\
&= \sum_{q=1}^Q \sum_{p=1}^P  \int  \normal(\varmeanwpq, \varcovwpq) \log \normal(0,\varw_{pq}) d w_{pq} \\
&= \sum_{q=1}^Q  \sum_{p=1}^P \left[\log \normal (\varmeanwpq ; \vec{0},\varcovwpq) - \frac{\varcovwpq}{2\varw_{pq}} \right]\text{,}
\end{align*}

where $\varcovwpq$ represents the $p$-th diagonal term of $\varcovwq$. 

 \section{Closed form evaluation of the ELL term}
 \label{sec:closed_form}

The \lgcpn model formulation allows to derive a closed form expression for the moments of the intensity function. Here we provide details about the derivations and obtain an expression for the first moment of $\exp(\wp\f_{n \bullet})$  which has been used in the closed form evaluation of $\ellterm$.  

In order to compute the moments of $\lambda$ we can exploit the moment generating function (\mgf) of the product of two normal random variables. Denote by $X$ and $Y$ two independent and normally distributed random variables. The variable $Z=XY$ has $\text{MGF}_Z(t)$ defined as:
\begin{equation}
\text{MGF}_Z(t) =  \frac{\text{exp}\left[\frac{t\mu_X\mu_Y + 1/2(\mu_Y^2\sigma_X^2 + \mu_X^2\sigma_Y^2)t^2}{1 - t^2 \sigma_X^2 \sigma_Y^2}\right]}{\sqrt{1-t^2\sigma_X^2\sigma_Y^2}}\text{.}
\label{eq:mgf_product}
\end{equation}

Now define $V =  \sum_{q=1}^{Q}X_{q}Y_q$ where $X_q \indep Y_q\text{,} \forall q$, $X_q \indep X_{q'}\text{,} \forall q, q'$ and $Y_q \indep Y_{q'}\text{,} \forall q, q'$. Given these assumptions, the \mgf for $V$ is defined as the product of $Q$ \mgf of the form given in \eq \eqref{eq:mgf_product}. We have $\text{MGF}_V(t) = \prod_{q=1}^{Q} \text{MGF}_{Z_q}(t)$. 
This implies that:
\begin{equation}
\mathbb {E}(\lambda_p) = \mathbb {E}\left[\text{exp}(\wp\f_{n \bullet}) \right]= \text{MGF}_V(1) 
\label{eq:mgf_1}
\end{equation}
where $X_q = \omega_{pq}$ and $Y_q = f_{nq}$.

Exploiting \eq \eqref{eq:mgf_1} we can derive a closed form expression for $\ellterm$: 
\begin{align*}
&\mathbb {E}_{q(\f), q(\W)}\left[\text{log}(p(\Y |  \f, \W))\right] =\\
&= -\sum_{n=1}^{N} \sum_{p=1}^{P} \mathbb {E}\left[\text{exp}(\wp\f_{n \bullet}+ \likeparam_p)  + y_{np}\text{log}(\text{exp}(\wp\f_{n \bullet} + \likeparam_p) \right. \\& 
\left.+ \text{log}\Gamma(y_{np} + 1)\right] \\
&= \sum_{n=1}^{N} \sum_{p=1}^{P} \mathbb {E}\left[-\text{exp}(\wp\f_{n \bullet} + \likeparam_p)  y_{np}\wp\f_{n \bullet} + y_{np}\likeparam_p \right. \\& 
\left.+ \text{log}\Gamma(y_{np} + 1)\right] \\
& =  -\sum_{n=1}^{N} \sum_{p=1}^{P} \mathbb{E} \left[\text{exp}(\wp\f_{n \bullet} + \likeparam_p)\right] + \sum_{n=1}^{N}  \sum_{p=1}^{P}\left[y_{np}\mathbb{E}(\wp\f_{n \bullet}) + \right. \\& 
\left. +y_{np}\likeparam_p + \text{log}\Gamma(y_{np} + 1)\right]\\
&=  -\sum_{n=1}^{N} \sum_{p=1}^{P} \text{exp}(\likeparam_p)MGF_{V}(1) + \\
&\sum_{n=1}^{N} \sum_{p=1}^{P} \sum_{q=1}^{Q}(y_{np} \omega_{pq} \mu_q(\x_n) + y_{np}\likeparam_p + \text{log}\Gamma(y_{np} + 1)) \numberthis
\label{lik}
\end{align*}

Given the moments of $q(\wp)$ and $q(\f_{n \bullet})$ we can write:
\begin{align*}
	&\mathbb {E}_{\qfn \qwp}\left[\text{exp}(\wp\f_{n \bullet}) \right] \\
	&= \prod_{q=1}^{Q}  \frac{\text{exp}\left[\frac{\omega_{pq}\mu_{nq}  + 1/2(\mu_{nq} ^2 \varcovwpq + \omega_{pq}^2  \varcovfq_{nn} )}{1 -  \varcovwpq \varcovfq_{nn}}\right]}{\sqrt{1-\varcovwpq \varcovfq_{nn}}} 
	\numberthis
	\label{eq:first_moment}
\end{align*}

Defining $\delta_X = \mu_{X}/ \sigma_X$ in \eq \eqref{eq:mgf_product} we can rewrite $\mgf_Z(t)$ as: 

\begin{equation}
\text{MGF}_Z(t) =  \frac{\text{exp}\left[\frac{t\mu_X\mu_Y + 1/2(\mu_Y^2\sigma_X^2 + \mu_X^2\sigma_Y^2)t^2}{1 - t^2 \frac{\mu_X^2 \sigma_Y^2}{\delta_X^2}}\right]}{\sqrt{1 - t^2 \frac{\mu_X^2\sigma_Y^2}{\delta_X^2}}}
\end{equation}

As $\delta_X$ increases, $\mgf_Z(t)$ converges to the form:
\begin{equation}
\text{MGF}_Z(t) =  \text{exp}\left[t\mu_X\mu_Y + 1/2(\mu_Y^2\sigma_X^2 + \mu_X^2\sigma_Y^2)t^2\right]
\end{equation}
which is the \mgf of a Gaussian distribution with mean and variance given by $\mu_X\mu_Y$ and $\mu_Y^2\sigma_X^2 + \mu_X^2\sigma_Y^2$ respectively \citep{seijas2012approach}. This implies that, for increasing values of $\delta_{X_q}$ the sum of the products of Gaussians tends to a Gaussian distribution.

\section{Relationship to existing literature}
As mentioned in \secref{sec:closed_form}, when $\varcovwq \to 0$, $\wp\f_{n \bullet}$ converges to a Gaussian distribution.  Depending on the number of latent \gptext{s} included in the model ($Q$) and the moments of $q(\wp)$, \lgcpn will thus converge either to an \icm (or \lcm) or to a \mlgcp or to an \lgcp. When $Q \neq P$, we have $\textnormal{log} \lambda_{p}(\xn) = \sum_{q=1}^{Q} \omega_{pq}\f_{n \bullet}$ for each $n$ and $p$. We can thus write:
 \begin{align*}
 %\label{eq:lemma}
 &\lim_{\covw \to 0} \textnormal{Cov}(\textnormal{log} \lambda_{p}(\x), \textnormal{log} \lambda_{p'}(\x')) \\
 &=
 \sum_{q=1}^{Q}  \sum_{q'=1}^{Q} \omega_{pq} \omega_{p'q'} \text{Cov}(\fq, \fqprime) \\
  &=\sum_{q=1}^{Q} \underbrace{\omega_{pq} \omega_{p'q'}}_{\vecS{B}_q} \covfuq_{\x\x'}
 \end{align*}
where we have exploited the independence assumption between $\fq$ and $\fqprime$ for $q \neq q'$.

When $Q=P+1$ and $\W_{P \times (P+1)} = [\vecS{I}_{P} \hspace{0.2cm} \mathds{1}_P]$, the intensity for each task will be determined by the $(P+1)$-th common \gptext and by the $p$-th task specific \gptext. We thus recover the \mlgcp formulation. 

Finally, when $Q=P$ and $\W_{P \times P} = \vecS{I}_{P}$, the intensity for each task will be determined only by the $p$-th task specific \gptext. We thus recover the \lgcp formulation. 

We summarise these results in the following lemma:
\begin{theorem} \label{lemma}
	\lgcpn generalizes \icm, \mlgcp and \lgcp.  As $\text{Cov}(w_{pq}, w_{p'q'}) \to 0 \text{,} \forall p,q,p',q'$, for $Q \neq P$ we have $\hat{\lambda}_{\lgcpn} \to \hat{\lambda}_{\icm}$ (or a $\hat{\lambda}_{\lgcpn} \to \hat{\lambda}_{\lcm}$ depending on the assumed covariance functions for the latent \gptext{s}) where the intensity parameters are jointly determined by the moments of $\f$ and  $\W$:
	\begin{align*}
	%\label{eq:lemma}
	\lim_{\substack{\text{Cov}(w_{pq}, w_{p'q'}) \to 0 \\ \forall p,q,p',q'}} &\textnormal{Cov}(\textnormal{log} \lambda_{p}(\x), \textnormal{log} \lambda_{p'}(\x')) 
	\\
	&=\sum_{q=1}^{Q} \underbrace{\priormeanwqp \priormeanwqprimepprime}_{\vecS{B}_{q_{(p,p')}}} \covfuq_{\x\x'}
	\end{align*}
	where $\vecS{B}_q \in \mathbb{R}^{P\times P}$ is known as coregionalisation matrix.
	For $Q=P+1$ and $\W_{P \times (P+1)} = [\vecS{I}_{P} \hspace{0.2cm} \mathds{1}_P]$ we have $\hat{\lambda}_{\lgcpn} \to \hat{\lambda}_{\mlgcp}$. Finally, for $Q=P$ and $\W_{P \times P} = \vecS{I}_{P}$ we have $\hat{\lambda}_{\lgcpn} \to \hat{\lambda}_{\lgcp}$. 
\end{theorem}

When having task descriptors $\h$, we can view the log intensity as a function of the joint space of input features and task descriptors $i.e.~\log \lambda(\x, \h)$. It is possible to show that under our independence prior assumption between weights $(\W)$ and latent functions $(\F)$, the prior covariance  over the log intensities (evaluated at inputs $\x$ and $\xprime$ and tasks $p$ and $\pprime$) is given by:
\begin{align*}
&\covariance{}{\log \lambdax} {\lambdaxprime}  
%&= 	\sum_q \sum_{\qprime}  \covariance{}{w_{q}(\hp)}{w_{\qprime}(\hpprime)} \covariance{}{f_q(\x)}{ f_{\qprime}(\xprime)} \\ 
&=\sum_{q=1}^Q \kwq(\hp, \hpprime) \kfq(\x, \xprime)
%\numberthis
%\label{eq:effective_prior}
\end{align*}
where $\hp$ denotes the $p$-th task descriptors. At the observed data $\{ \X, \H \}$,  assuming a regular  grid, the \lgcpn prior covariance over the log intensities is $\covariance{}{\log \pmb{\lambda}(\X)}{ \log \pmb{\lambda}(\X)} = \sum_{q=1}^{Q} \Kwq \kron \Kfq$. This is effectively the \lcm prior  with  $\Kwq$ denoting  the coregionalization matrix. 
Importantly, the two methods differ substantially in terms of inference. While in \lcm a point estimate of $\Kwq$ is generally obtained, \lgcpn proceeds by optimizing the hyperparameters for $\Kwq$ and doing full posterior estimation for both $\W$ and $\f$. In addition, by adopting a process view on $\W$, we increase the model flexibility and accuracy by capturing additional correlations across tasks while being able to generalize over unseen task descriptors. Last but not least, by considering our priors and approximate posteriors over $\W$ and $\F$ separately, instead of a single joint prior over the log intensities, we can exploit state-of-the art inducing variable approximations \citep{2009variational} over each $\wq$ and $\fq$ separately, instead of dealing with a sum of $Q$ Kronecker products for which there is not an efficient decomposition when $Q > 2$ \citep{rakitsch2013all}.  

\section{Continuous \lgcpn formulation}

Following a common approach, in this work we introduce a computational grid on the spatial extend and consider the cells' centroids as inputs of \lgcpn. Here we discuss the continuous formulation of our model. The likelihood function for the continuous \lgcpn model can be written as:
\begin{align*}
p(Y|\lambda) = \text{exp}\left[- \sum_{p=1}^{P}\int_{\tau}\lambda_p(\x)dx\right] \prod_{p=1}^{P} \prod_{n_p }^{N_p} \lambda_p(\x_{n_p})
\end{align*}
where we assume all events to be distinct and we denote as $n_p$ the location of the $n$-th event for the $p$-th task. This implies an expected log likelihood term defined as:
\begin{align*}
&\mathbb{E}_{q(\f)q(\W)}\left[-\sum_{p}^{P}\int_{\tau}\lambda_p(\x)d\x + \sum_{p=1}^{P}\sum_{n_p}^{N_p} \text{log}\lambda_p(\x_{n_p})\right]
\end{align*}

Replacing the expression for \lgcpn intensity in the previous equation we get:
\begin{align*}
 &\mathbb{E}_{q(\f)q(\W)}\text{log}(p(Y|\lambda)) \\
 &= -\sum_{p=1}^{P}\int_{\tau}\int_{\f}\int_{\W}\text{exp}(\wp\f_{n_p \bullet})q(\W)q(\f)d\W d\f  d\x + \\
&+ \sum_{p=1}^{P} \sum_{n_p}^{N_p} \int_{\tau}\int_{\f}\int_{\W}\text{log}\left[\text{exp}(\wp\f_{n_p \bullet})\right]q(\W)q(\f)d\W d\f  \\
&= -\sum_{p=1}^{P} \int_{\tau}\mathbb{E}\left[\text{exp}(\wp\f_{n_p \bullet})\right]d\x \\
&+ \sum_{p=1}^{P} \sum_{n_p}^{N_p}\int_{\tau}\int_{\f}\int_{\W}\wp\f_{n_p \bullet}q(\W)q(\f)d\W d\f  d\x \\
&= -\sum_{p=1}^{P} \int_{\tau}\mathbb{E}\left[\text{exp}(\wp\f_{n_p \bullet})\right]d\x + \sum_{p=1}^{P} \sum_{n_p}^{N_p}\mathbb{E}(\wp\f_{n_p \bullet})
\end{align*}
for a  bounded region $\tau$. The expected value of $\text{exp}(\wp\f_{n \bullet})$ can be computed as in \eq \ref{eq:first_moment} while $\mathbb{E}_{q(\f)q(\W)}(\sum_{q=1}^{Q}w_pf(\x_{n_p}))$ is equal to:
\begin{align}
\mathbb{E}_{q(\f)q(\W)}(\sum_{q=1}^{Q}w_pf(\x_{n_p}))  = \sum_{q=1}^{Q}\omega_p\mu_q(\x_{n_p}))
\end{align}

We are thus left with an intractabe integral of the form:

\begin{align*}
- \sum_{p=1}^{P} &\left[\prod_{q=1}^{Q} \frac{1}{\sqrt[]{1-\varcovwpq^2 \varcovfq_{nn}}} \text{exp}\left(- \frac{\omega_{pq}^2}{2\varcovwpq^2} \right) ... \right. \\& 
\left. ... \int_{\tau}\text{exp}\left(\frac{(\varcovwpq^2\varmeanfqn(\x) + \omega_{pq})^2}{\varcovwpq^2 \varcovfq_{nn}-1}\right)d\x\right]
\end{align*}
where the posterior mean for $q(\f)$ computed in $\x$ is defined as $\varmeanfqn(\x) = k_{x z}^q(K_{zz})^{-1}m_q$.

This integral could be approximated using a series expansion but this would result in a computationally difficult problem.
\section{Pseudo-algorithm}
Algorithm 1 illustrates the \lgcpn algorithm:
\begin{algorithm}[H]
\caption{LGCPN}\label{LGCPN_code}
\begin{algorithmic}[1]
\Inputs{} Observational dataset $\mathcal{D} = \{\x^{(i)}_p \in \tau, \forall p =1,...,P\}_{i=1}^I$ for bounded region $\tau$ where $I$ denotes the number of events. Number of latent \gptext{s} $Q$. Number of mini-batches $\texttt{b}$ of size $B$.
\Output{} Optimized hyper-parameters, posterior moments of $\vecS{\lambda}$ \\

\prediction{}Discretize event locations $\mathcal{D}$ in $Y \in \mathbb{R}^{N \times P}$ given the grid size. 
\Initialize{}  $i \gets 0$, $\allparameters^{(0)} = (\hyperparam, \hyperparam_w, \likeparamvector, \varparam_u, \varparam_w)$
%$\priormeanw \gets 0$, $\priormeanu \gets 0$
\Repeat
\State $\{X_{train} \in \mathbb{R}^{B\times D}$, $Y_{train} \in \mathbb{R}^{B\times P}\} \to \texttt{get-next-MiniBatch}(\mathcal{D})$
\For{\texttt{j=0 to b}}
\State $\max_{\pmb{\mu}} \elbo(\allparameters^{(i)})$ (\eqs \eqref{eq:ent_u}--\eqref{eq:crossent_w} and \eqref{lik})
\State $\allparameters^{(i)} \gets \allparameters^{(i-1)}- \rho \nabla_{\allparameters} \elbo(\allparameters^{(i-1)})$
\State $i = i +1$
\EndFor
\Until{convergence criterion is met.}
\prediction{} $\allparameters^* \gets \allparameters^{(i - 1)}$
\prediction{} $\mathbb{E}\left[\vecS{\lambda}(\x)^t\right] = \text{exp}(t\likeparamvector^*) \mgf_{\W\f|\allparameters^*}(t)$
\end{algorithmic}
\end{algorithm}

%\begin{wrapfigure}[19]{R}{0.45\textwidth}
%	\vspace{-.3cm}
%	\centering
%	\includegraphics[width=.9\linewidth]{figs/plate_asfigure}
%	\caption{\lgcpn - Plate diagram. $X'_{pd'}$ represents the inputs for the \gptext prior on $\W$. When placing a Normal prior on each $w_{pq}$, we introduce the additional factorization across $P$ (dashed plate). }
%	\label{fig:plates}
%\end{wrapfigure}
\section{Plate diagram}
Here we provide the plate diagram for \lgcpn with independent prior on the mixing weights:
\begin{figure}[h]
	\begin{center}
	%\vspace{-.3cm}
	\includegraphics[width=6.cm,height=5.90cm,  valign = c]{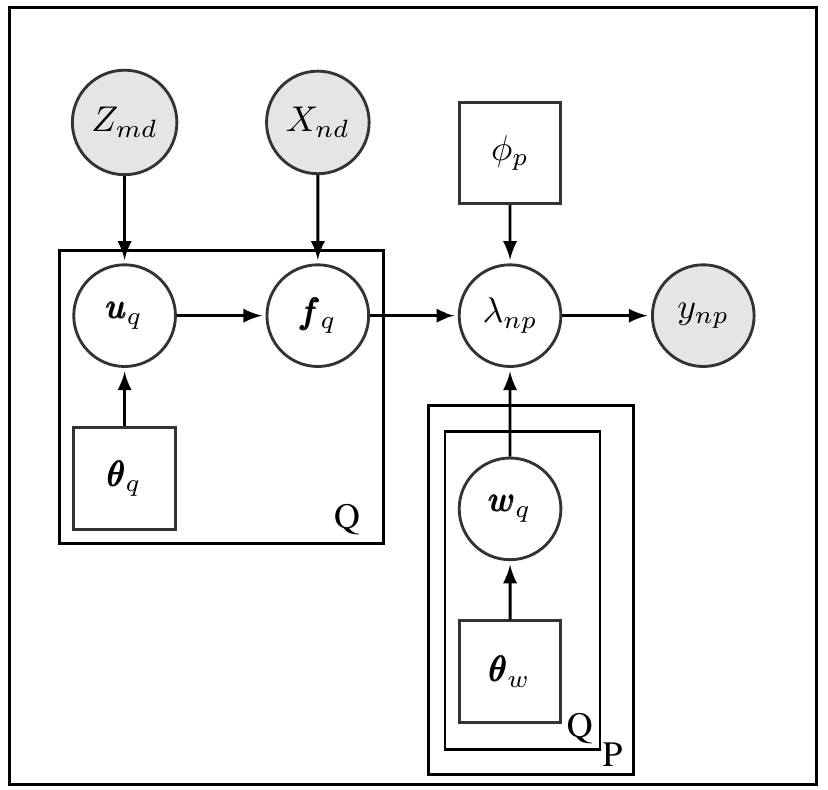}
	\caption{Graphical model representation of \lgcpnnormal.}
	\label{fig:plates}
\end{center}
\end{figure}

% Efficiency discussion 
\section{Algorithmic efficiency}
\begin{figure}[H]
	\begin{center}
		\includegraphics[width=6.cm,height=3.90cm, valign = c]{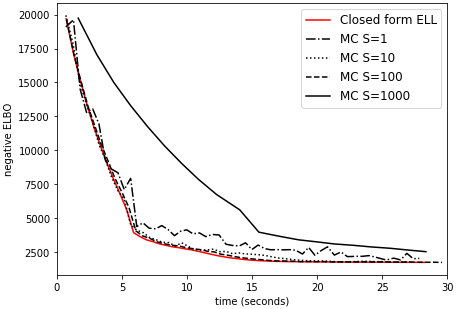}
		\vspace{0.1cm}
		\includegraphics[width=6.cm,height=3.92cm, valign = c]{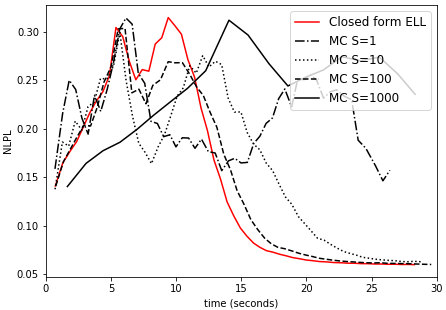}
		\caption{Synthetic data. MC estimate of ELL vs. Closed form evaluation of ELL. \textit{Left:} Negative \termelbo values over time. \textit{Right:} \nlpl values for one task over time. $S$ denotes the number of samples used in the MC evaluation.}
		\label{fig:time_elbos_synthetic}
	\end{center}
\end{figure}

\begin{figure}[H]
	\begin{center}
		\includegraphics[width=6.cm,height=3.90cm, valign = c]{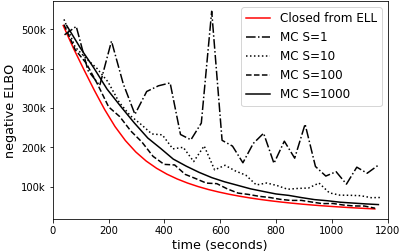}
		\vspace{0.1cm}
		\includegraphics[width=6.cm,height=3.92cm, valign = c]{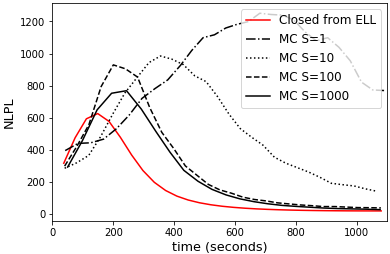}
		\caption{\crime data. MC estimate of ELL vs. Closed form evaluation of ELL. \textit{Left:} Negative \termelbo values over time. \textit{Right:} \nlpl values for one task over time. $S$ denotes the number of samples used in the MC evaluation.}
		\label{fig:time_elbos_crime}
	\end{center}
\end{figure}

Evaluating $\ellterm$ in closed form, we are able to significanlty speed up the algorithm by getting rid of the Monte Carlo evaluations, see \fig \ref{fig:time_elbos_synthetic} and \fig \ref{fig:time_elbos_crime}.

% Figures form time - elbo -preformance

\section{Additional experimental results}
\begin{figure*}
	\begin{center}
		\includegraphics[width=0.8\textwidth]{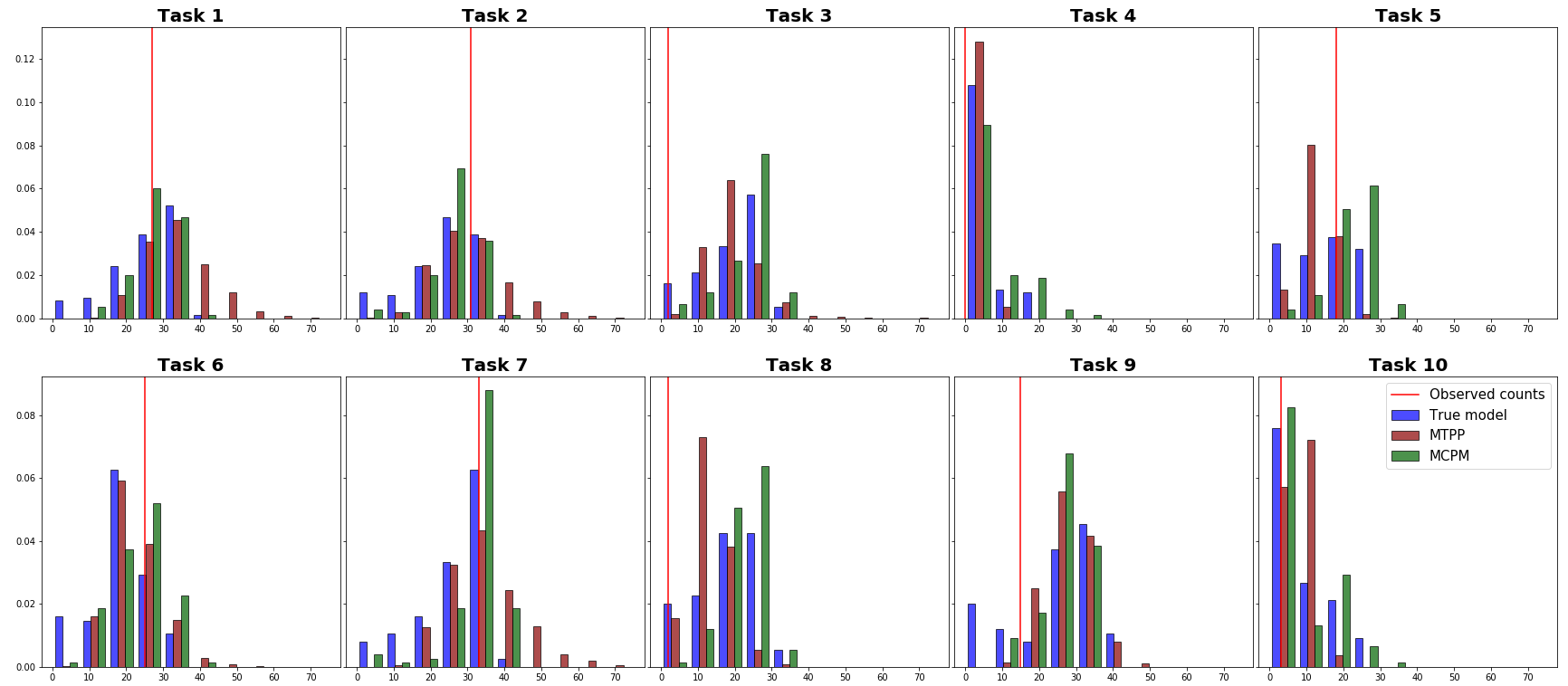}
		\caption{Predicted empirical distributions of event counts in $[80,100]$ for the second synthetic dataset.}
		\label{fig:predicted_hist_mtpp_all}
	\end{center}
\end{figure*}

\paragraph{Synthetic experiments}
Here we report additional performance metrics for the two synthetic experiments included in the text. \tbl \ref{tab:coverage_synthetic1} gives the coverage numbers for the first synthetic experiment while \tbl \ref{tab:rmse_synthetic2} and \tbl \ref{tab:coverage_synthetic2} display the \rmse and coverage performances for the second synthetic dataset. \fig \ref{fig:predicted_hist_mtpp_all} gives the predicted counts distributions for the second synthetic dataset.

\begin{table}[h]
		\caption{\sinone dataset. In-sample/Out-of-sample 90\% \ci coverage for the predicted event counts distributions.}
	\begin{center}
		\resizebox{0.5\textwidth}{!}{\begin{tabular}{lccccc}
				\toprule
				\multirow{3}{*}{} &
				\multicolumn{4}{c}{Empirical Coverage (\coverage) } \\
				\cmidrule(r){2-5}
				& \textbf{1} & \textbf{2} & \textbf{3} & \textbf{4} \\
				\midrule
				{\lgcpnnormal} & 0.80/0.12 	& \textbf{0.99}/0.58 & 0.92/0.57 & \textbf{0.94}/\textbf{0.83} \\  
				{\lgcpngp} & \textbf{0.95}/\textbf{0.19} & 0.72/\textbf{0.67} & \textbf{1.00}/\textbf{0.78} & 0.92/0.75\\
				%{\lgcp} & 0.95/0.14 & 1.00/0.60 & 0.84/0.10 & 0.99/0.92\\
				{\icm}  & 0.75/0.03	& 0.66/0.60 & 0.62/0.50 &0.93/0.42 \\
				
				\bottomrule
		\end{tabular}}
	\end{center}
	\label{tab:coverage_synthetic1} 
\end{table}

\begin{table}[H]
	\caption{\sintwo dataset. \rmse performance when making predictions on the interval $[80,100]$.}
	\label{tab:rmse_synthetic2}
	\begin{center}
		\resizebox{0.49\textwidth}{!}{\begin{tabular}{lccccccccccc}
				\toprule
				\multirow{3}{*}{} &
				\multicolumn{10}{c}{\rmse } \\
				\cmidrule(r){2-11}
				& \textbf{1} & \textbf{2} & \textbf{3} & \textbf{4} & \textbf{5} & \textbf{6} & \textbf{7} & \textbf{8} & \textbf{9} & \textbf{10}\\
				\midrule
				{\lgcpnnormal} & \textbf{1.10} & \textbf{1.15} & \textbf{0.89} & 0.17 & 0.95 &
				0.99 & 1.10 & 0.63 & 1.50 & 0.55  \\
				{\lgcpngp} & 1.15 & 1.43 & 0.91 & \textbf{0.13} & 0.94& \textbf{0.97} &1.19& \textbf{0.58} & \textbf{1.43} & 0.70\\
				{\mtpp} & 1.20 & 1.70 & 1.12 & 0.17 & \textbf{0.91} &
				1.05 & \textbf{1.05} & 1.11 & 1.61 & \textbf{0.49} \\
				
				\bottomrule
		\end{tabular}}
	\end{center}
\end{table}

\begin{table}[H]
		\caption{\sintwo dataset. In-sample/Out-of-sample 90\% \ci coverage for the predicted event counts distributions.}
	\begin{center}
		\resizebox{0.5\textwidth}{!}{\begin{tabular}{lccccccccccc}
				\toprule
				\multirow{3}{*}{} &
				\multicolumn{10}{c}{Empirical Coverage } \\
				\cmidrule(r){2-11}
				
				& \textbf{1} & \textbf{2} & \textbf{3} & \textbf{4} & \textbf{5} & \textbf{6} & \textbf{7} & \textbf{8} & \textbf{9} & \textbf{10}\\
				\midrule
				{\lgcpnnormal} & \textbf{1.00}/\textbf{1.00}  & \textbf{1.00}/\textbf{1.00}  & 0.95/0.99 & 0.66/\textbf{1.00}  & \textbf{1.00}/0.86 & 0.97/\textbf{1.00} & \textbf{0.99}/\textbf{1.00 } & 0.88/\textbf{1.00}  & 0.92/0.95 & \textbf{1.00}/\textbf{1.00} &  \\
				{\lgcpngp} & 0.99/\textbf{1.00} & \textbf{1.00}/\textbf{1.00}  & \textbf{1.00}/\textbf{1.00}  & \textbf{1.00}/\textbf{1.00}  & \textbf{1.00}/\textbf{1.00}  & \textbf{1.00}/\textbf{1.00} & \textbf{0.99}/\textbf{1.00} & \textbf{1.00}/\textbf{1.00}  & \textbf{1.00}/\textbf{1.00}  & \textbf{1.00}/\textbf{1.00}  \\
				{\mtpp} &  0.77/0.77 &   0.82/0.73  &  0.86/\textbf{1.00}  &  0.93/\textbf{1.00}  &  0.75/0.83  & 0.96/0.84   & 0.78/0.54 &  0.99/\textbf{1.00} & 0.66/0.88  & 0.74/0.95 \\
				
				\bottomrule
		\end{tabular}}
	\end{center}
	 \label{tab:coverage_synthetic2}
\end{table}

% plots for mtpp

\subsection{Crime data experiments}  
\begin{figure}
	\begin{center}
		\includegraphics[width=0.5\textwidth]{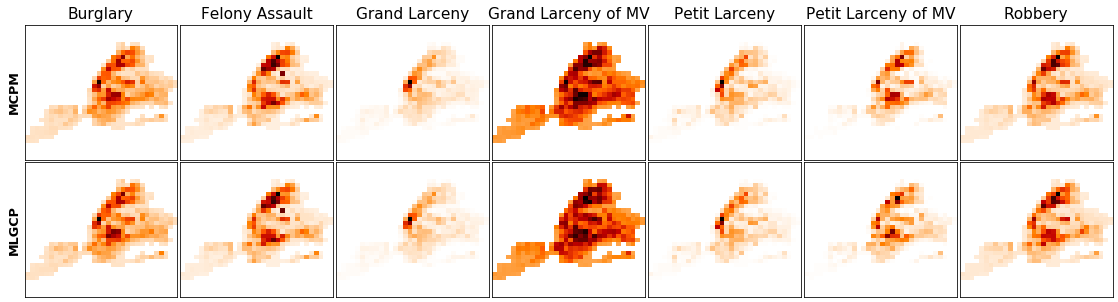}
		\caption{\label{fig:LGCPN_onCrime}\crime dataset. Estimated intensity surface with \lgcpn (\textit{first row}) and \mlgcp (\textit{second row}). The color scale used for each crime is given in \fig (5).} \label{fig:LGCPN_onCrime_missing_int}
		\end{center}
\end{figure}

\begin{figure}
	\begin{center}
		\centerline{\includegraphics[width=0.5\textwidth]{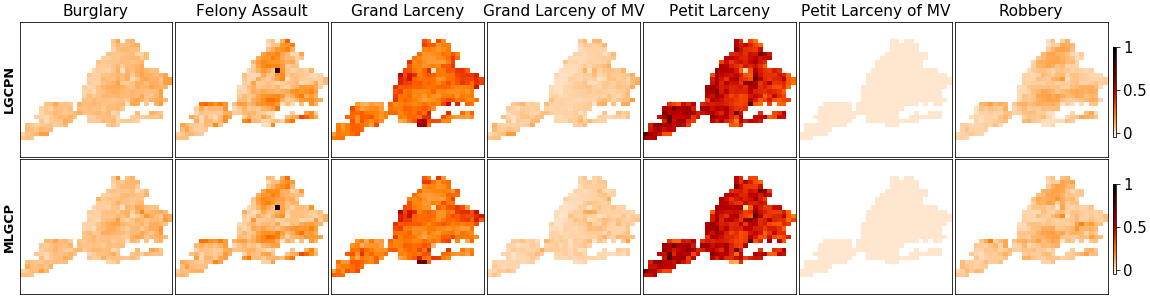}}
		\caption{\crime dataset. Estimated conditional probabilities in the complete data setting. \textit{Row 1:} \lgcpn \textit{Row 2:} \mlgcp. }
		\label{fig:LGCPN_onCrime_missing_condprob}
	\end{center}
\end{figure}

\begin{figure}
	\begin{center}
		\centerline{\includegraphics[width=0.5\textwidth]{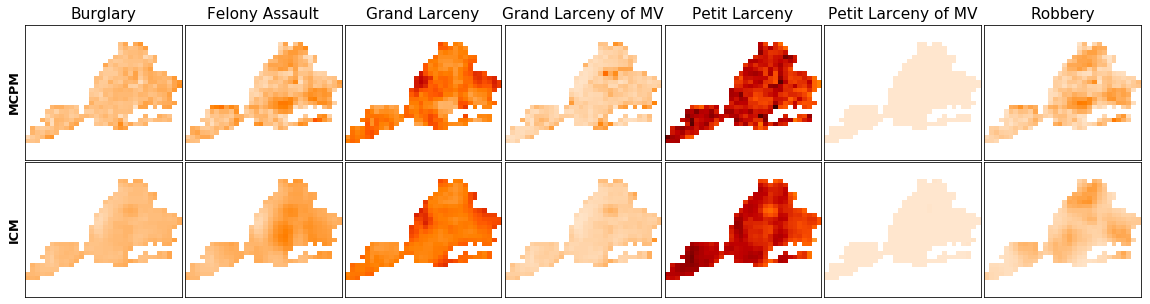}}
		\caption{\crime dataset. Estimated conditional probabilities when introducing missing data regions. \textit{Row 1:} \lgcpn \textit{Row 2:} \lgcp. }
		\label{fig:LGCPN_onCrime_missing_condprob_icm}
	\end{center}
\end{figure}

Here we report the \rmse performances for \lgcpn and competing models on the \crime dataset (\tbl \ref{tab:performance_crime_missing}). In \fig \ref{fig:LGCPN_onCrime_missing_int} and \fig \ref{fig:LGCPN_onCrime_missing_condprob} we give the estimated intensities and conditional probabilities for the \crime complete data experiment. Finally, in \fig \ref{fig:LGCPN_onCrime_missing_condprob_icm} we show the conditional probabilities for the missing data experiment.
\begin{table}
	\caption{\crime dataset. Performance on the missing regions. Standard errors in parentheses.}
	% These are the results on the land cells
	\label{tab:performance_crime_missing}
	\begin{center}
		\resizebox{0.48\textwidth}{!}{\begin{tabular}{lccccccccccccccc}
				\toprule
				\multirow{2}{*}{} &
				\multicolumn{7}{c}{Standardized \rmse} \\
				\cmidrule(r){2-8}
				& \textbf{1} & \textbf{2} & \textbf{3} & \textbf{4} & \textbf{5} & \textbf{6} & \textbf{7} \\
				\midrule
				
				\multirow{2}{*}{\lgcpn}&  1.74 &   2.91 &  \textbf{3.00} &   \textbf{2.75} &
				3.57 &  \textbf{11.70} & \textbf{1.54} \\
				&(0.42) &  (1.06) &  (1.22) &  (0.82) &  (1.99) &
				(2.32) &  (0.29)  \\
				
				\multirow{2}{*}{\lgcpngp}  & \textbf{1.71} &  \textbf{1.91} &   3.40  &   2.96 &
				\textbf{2.00} &  12.18 &   1.62  \\
				&(0.39) &  (0.33) &  (1.80) &(1.03) & (0.47) &
				(2.76) & (0.33)  \\
				
				\multirow{2}{*}{\lgcp} &   5.16 &  4.68 &   8.93 &   3.09 &
				7.69 &  36.96 & 5.19 \\
				& (1.81) &  (0.99) & (5.22) & (0.50) & (3.68) &
				(5.43) &  (1.21) \\
				
				\multirow{2}{*}{\icm} &  3.36 &   3.64 &  3.70 &   2.97 &
				3.05 &  12.36 & 2.82 \\
				& (1.04) &  (0.83) &  (1.89) &  (1.22) &  (0.97) &
				(1.99) &  (0.62)  \\

				\bottomrule
		\end{tabular}}
	\end{center}
\end{table}

\subsection{BTB data experiments} 
\begin{figure}
	\begin{center}
		\centerline{\includegraphics[width=0.12\textwidth]{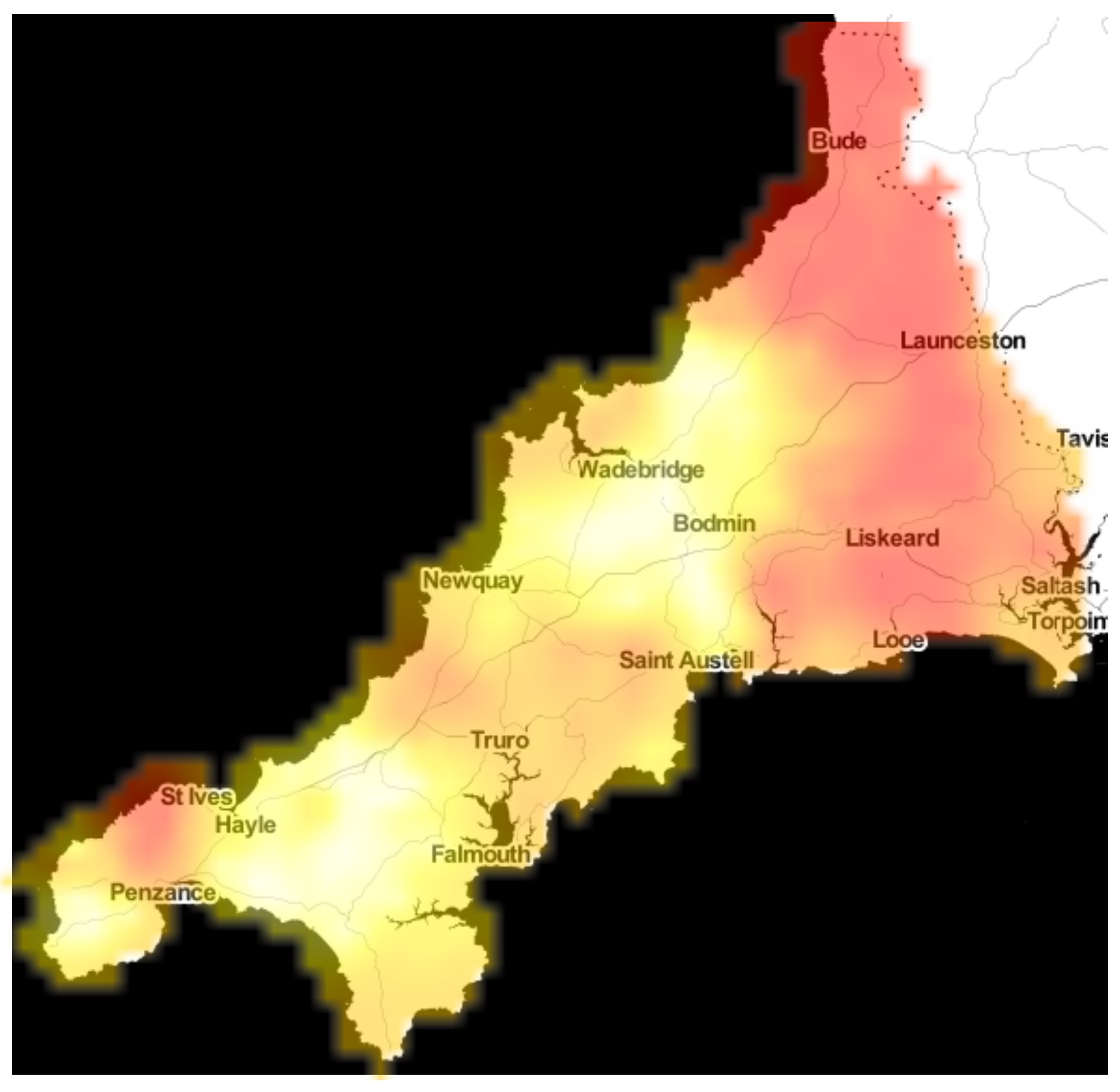}
			\includegraphics[width=0.12\textwidth]{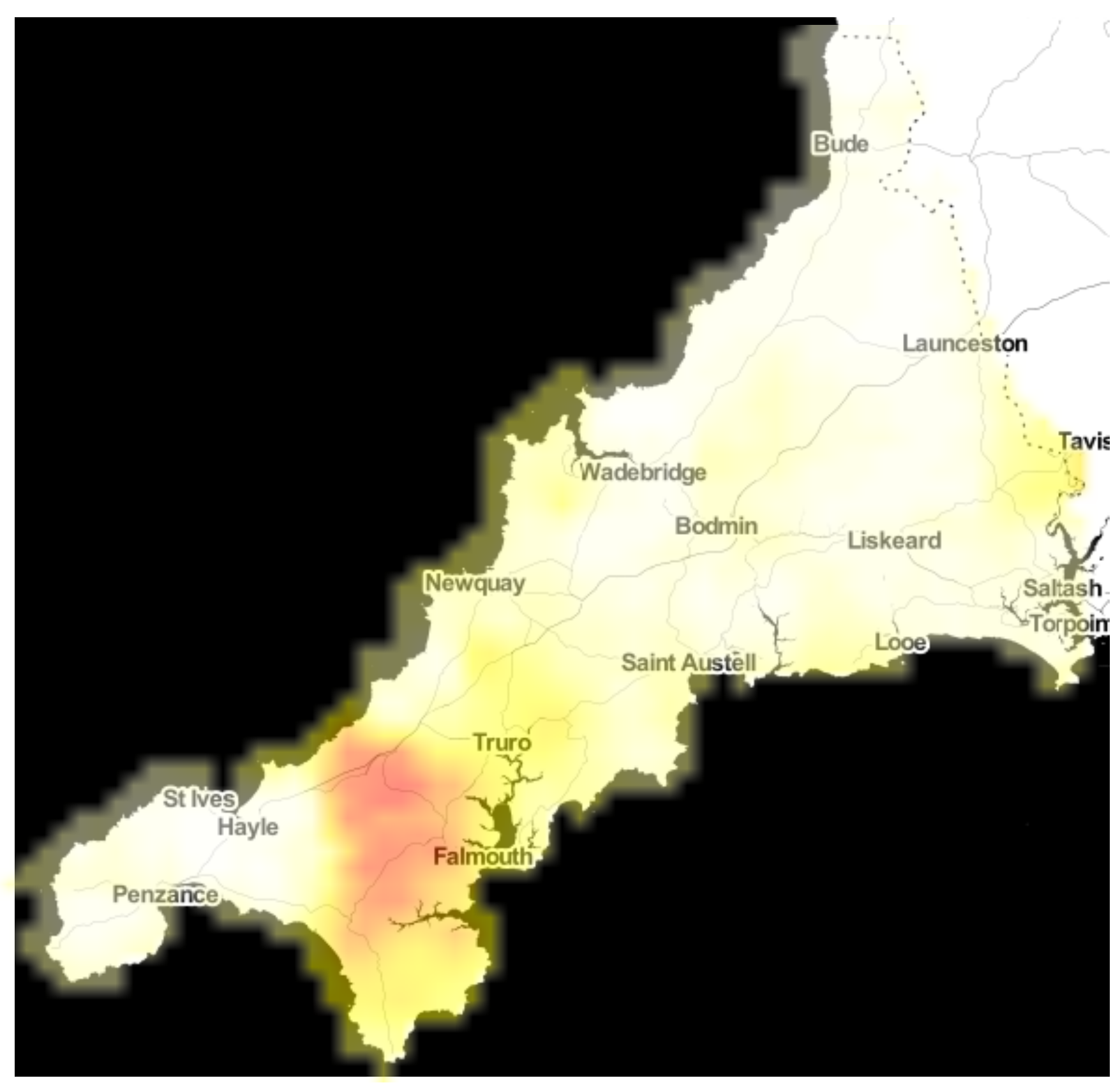}
			\includegraphics[width=0.12\textwidth]{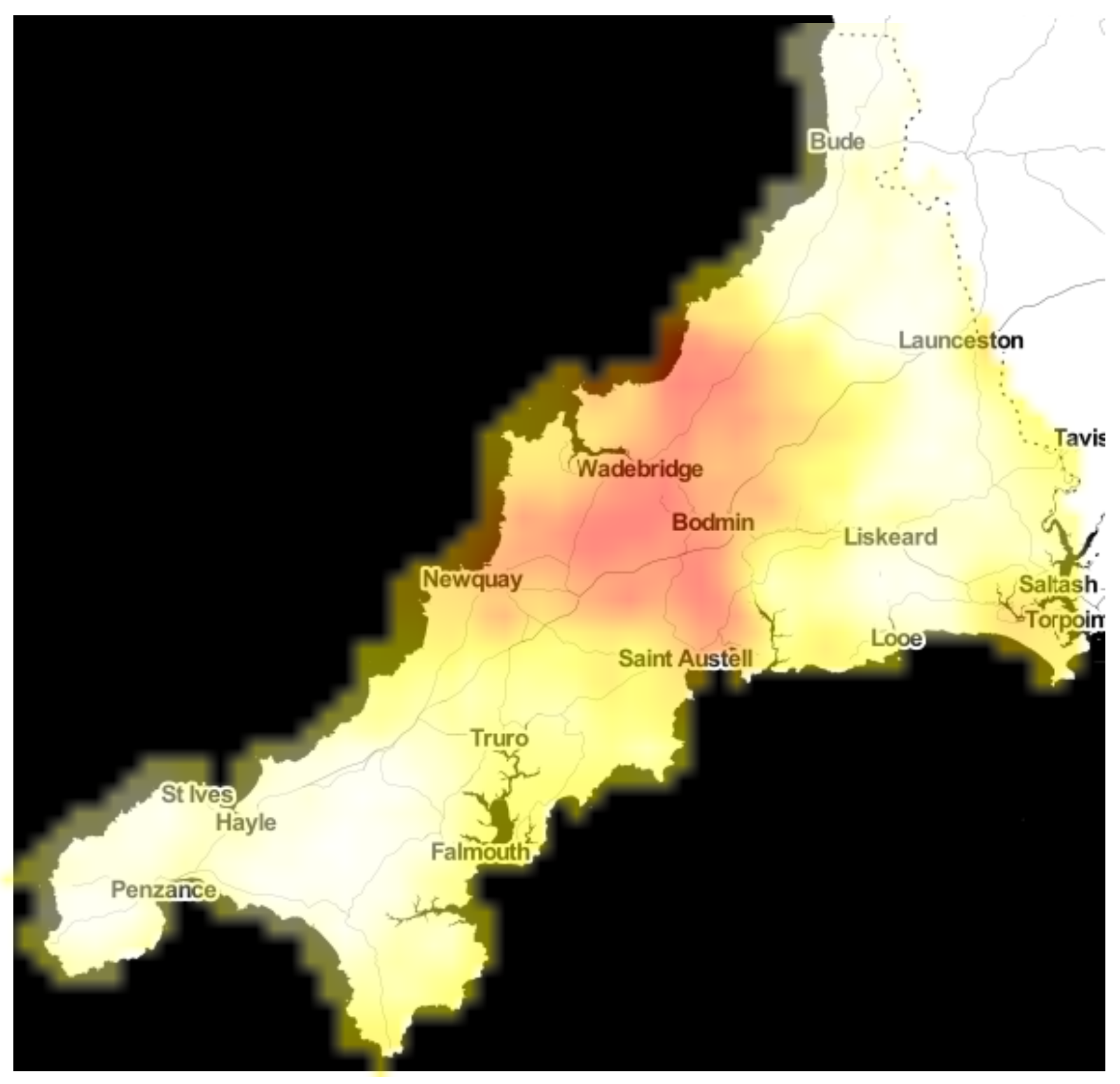}
			\includegraphics[width=0.12\textwidth]{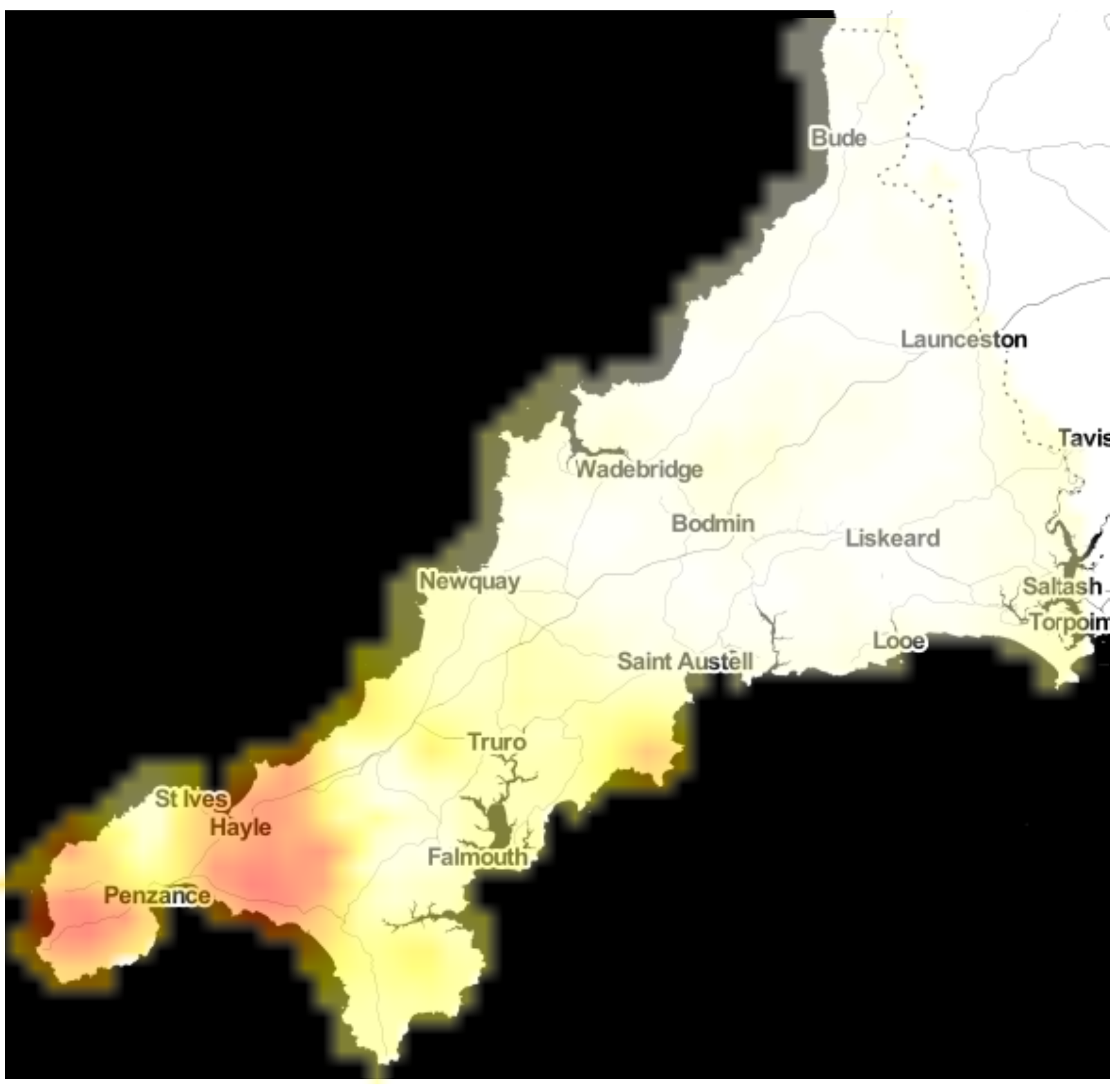}}
		\caption{\mlgcp - \btb dataset. Estimated conditional probabilities plotted on the color scale used by \citet{diggle2013spatial} and \citet{taylor2015bayesian}. The first plots corresponds to \gt 9, the second to \gt 12, the third to \gt 15 and the fourth to \gt 20.}
		\label{fig:mLGCP_original_scale}
	\end{center}
\end{figure}

\begin{figure}
	\begin{center}
\includegraphics[width=0.46\textwidth]{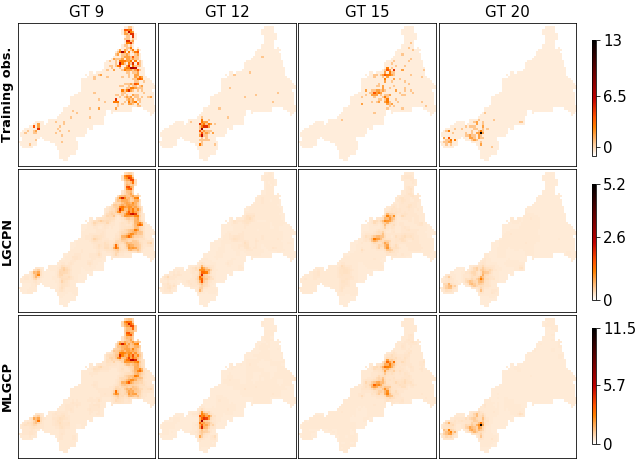}
\captionof{figure}{Estimated intensity surfaces in the complete data setting. \textit{First row:} Training data. \textit{Second row:} \lgcpn \textit{Third row:} \mlgcp}
\label{fig:surfaces_complete_BTB}
	\end{center}
\end{figure}

\begin{figure}
	\begin{center}
			\includegraphics[width=0.46\textwidth]{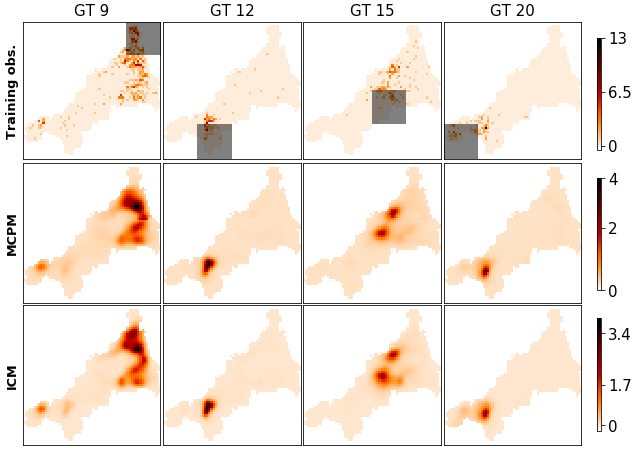}
	\captionof{figure}{Estimated intensity surfaces in the missing data (shaded regions) setting. \textit{First row:} Training data. \textit{Second row:} \lgcpn \textit{Third row:} \icm}
	\label{fig:surfaces_transfer_BTB}
	\end{center}
\end{figure}

In \fig \ref{fig:mLGCP_original_scale} we show the estimated conditional probabilities on the origin color scale used by \cite{diggle2013spatial}. In \fig \ref{fig:surfaces_complete_BTB} we give the estimated intensity surfaces for the complete data experiment. Finally, in \fig \ref{fig:surfaces_transfer_BTB} we show the estimated intensity surfaces for the missing data experiment.

\bibliographystyle{apalike}
\bibliography{references}